\documentclass[10pt,twocolumn,letterpaper]{article}

\usepackage[pagenumbers]{cvpr} 

\usepackage{bm}
\usepackage{graphicx}
\usepackage{amsmath}
\usepackage{amssymb}
\usepackage{booktabs}
\usepackage{mathtools}
\usepackage{booktabs}
\usepackage{multirow}
\usepackage[table]{xcolor}
\definecolor{lightgray}{gray}{0.9}
\usepackage[pagebackref,breaklinks,colorlinks]{hyperref}
\usepackage[accsupp]{axessibility} 

\usepackage[capitalize]{cleveref}
\crefname{section}{Sec.}{Secs.}
\Crefname{section}{Section}{Sections}
\Crefname{table}{Table}{Tables}
\crefname{table}{Tab.}{Tabs.}

\def\cvprPaperID{2234} 
\def\confName{CVPR}
\def\confYear{2023}

\def\forexample{\emph{e.g}\onedot}
\def\thatis{\emph{i.e}\onedot}

\def\secmk{Sec.~}
\def\figmk{Fig.~}
\def\tablemk{Tab.~}
\def\equationmk{Eqn.~}
\def\supmat{{\color{red}{Sup.~Mat~}}}

\def\Equal{\texttt{=}}
\def\shortminus{%
  \setbox0=\hbox{-}%
  \vcenter{%
    \hrule width\wd0 height \the\fontdimen8\textfont3%
  }%
}
\newcommand*{\mybox}[1]{\framebox{#1}}

\def\mocap{MoCap~}
\def\datanum{100K~}
\def\splicevit{Splice-ViT~}
\def\dinovit{DINO-ViT~}

\begin{document}
\title{Semi-supervised Hand Appearance Recovery via Structure Disentanglement and Dual Adversarial Discrimination}

\author{
  Zimeng Zhao\hspace{2mm}\hspace{5mm} 
  Binghui Zuo\hspace{2mm}\hspace{5mm} 
  Zhiyu Long\hspace{2mm}\hspace{5mm} 
Yangang Wang\footnotemark[1]\\%
\\
Southeast University, China\\
}

\maketitle
\begin{abstract}
   Enormous hand images with reliable annotations are collected through marker-based \mocap. Unfortunately, degradations caused by markers limit their application in hand appearance reconstruction. A clear appearance recovery insight is an image-to-image translation trained with unpaired data. However, most frameworks fail because there exists structure inconsistency from a degraded hand to a bare one. 
   The core of our approach is to first disentangle the bare hand structure from those degraded images and then wrap the appearance to this structure with a dual adversarial discrimination (DAD) scheme. 
   Both modules take full advantage of the semi-supervised learning paradigm: The structure disentanglement benefits from the modeling ability of ViT, and the translator is enhanced by the dual discrimination on both translation processes and translation results. 
   Comprehensive evaluations have been conducted to prove that our framework can robustly recover photo-realistic hand appearance from diverse marker-contained and even object-occluded datasets. It provides a novel avenue to acquire bare hand appearance data for other downstream learning problems. 
\end{abstract}

\renewcommand{\thefootnote}{\fnsymbol{footnote}}
\footnotetext[1]{Corresponding author. E-mail: yangangwang@seu.edu.cn. This work was supported in part by the National Natural Science Foundation of China (No. 62076061), in part by the Natural Science Foundation of Jiangsu Province (No. BK20220127).}

\section{Introduction} 
Both bare hand appearance and vivid hand motion are of great significance for virtual human creation. A dilemma hinders the synchronous acquisition of these two: accurate motion capture~\cite{han2018online, glauser2019interactive,sundaram2019learning} relies on markers that degrade hand appearance, whereas detailed appearance capture~\cite{wang2019hand,qian2020html,moon2020deephandmesh} in a markerless setting makes hand motion hard to track. Is there a win-win solution that guarantees high fidelity for both? 

\begin{figure}[!t]
    \centering
    \includegraphics[width=\linewidth]{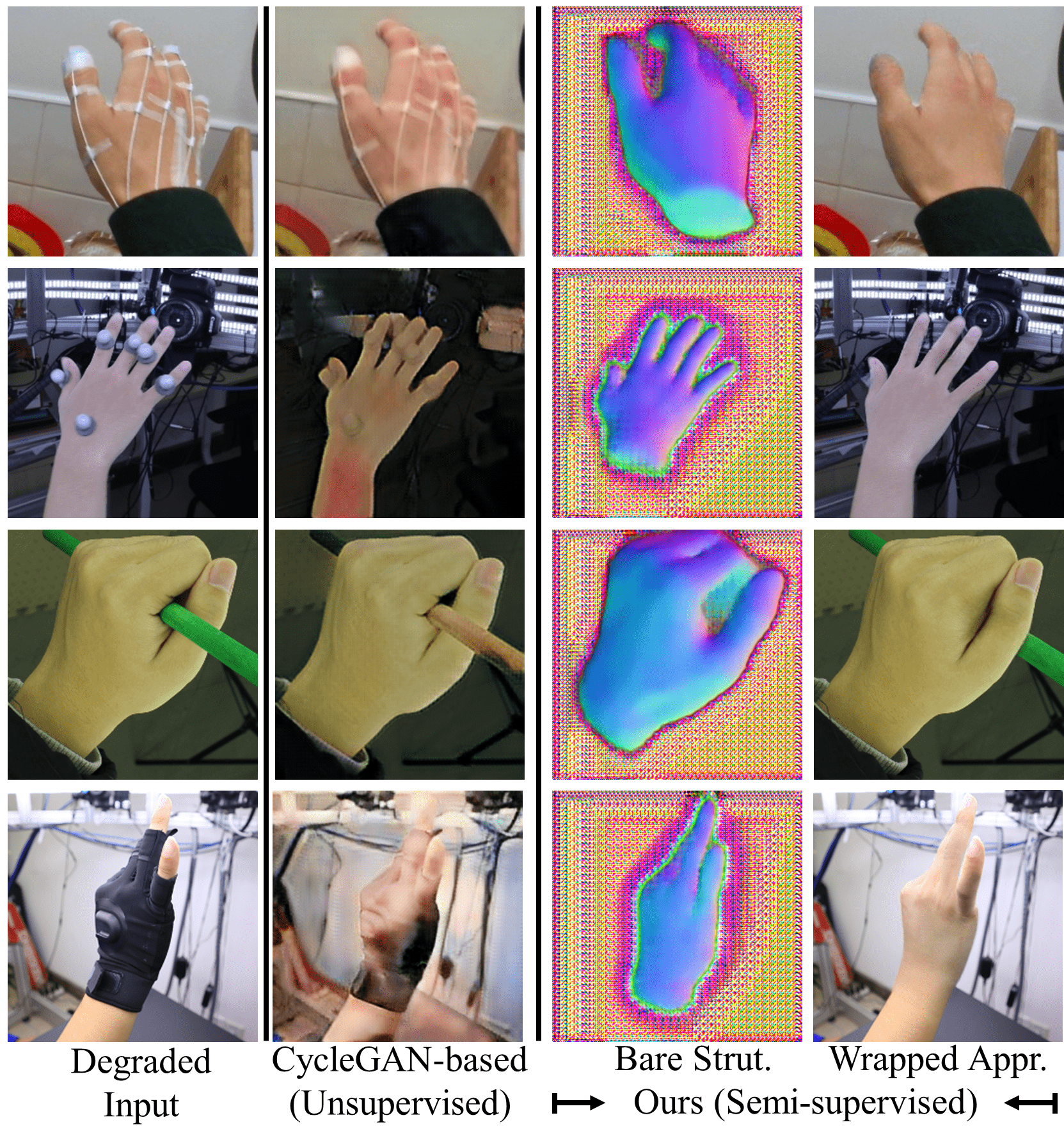}
    \vspace{-8mm}
    \caption{\textbf{Hand appearance recovery from diverse degradations}. Compared with CycleGAN-based frameworks, we recover more \emph{bare} hand appearance while preserving more semantics.}
    \vspace{-4mm}
    \label{teaser_optRes}
\end{figure}

Existing ones include markerless \mocap\cite{zhang20163d, han2020megatrack, zhao2020hand} and graphic rendering~\cite{hasson2019learning, yang2020seqhand, gao2022dart}. However, the former requires a pose estimator~\cite{lv2021handtailor, zhou2020monocular, chen2021camera} trained with laborious annotations. And the latter often produces artifacts because 
it is hard to simulate photo-realistic lighting. Another insight is to ``translate'' the degraded appearances as bare ones end-to-end. Nevertheless, it is tough to collect paired data for its training. Moreover, most unsupervised frameworks~\cite{zhu2017unpaired,park2020contrastive,oprea2021h} are only feasible when the translating target and source are consistent in structure, while our task needs to change those marker-related structures in the source. To this end, our key idea is to \textbf{first disentangle the bare hand structure represented by a pixel-aligned map, and then wrap the appearance on this bare one trained with a dual adversarial discrimination (DAD) scheme}. 

\begin{figure*}[!t]
    \centering
    \includegraphics[width=\linewidth]{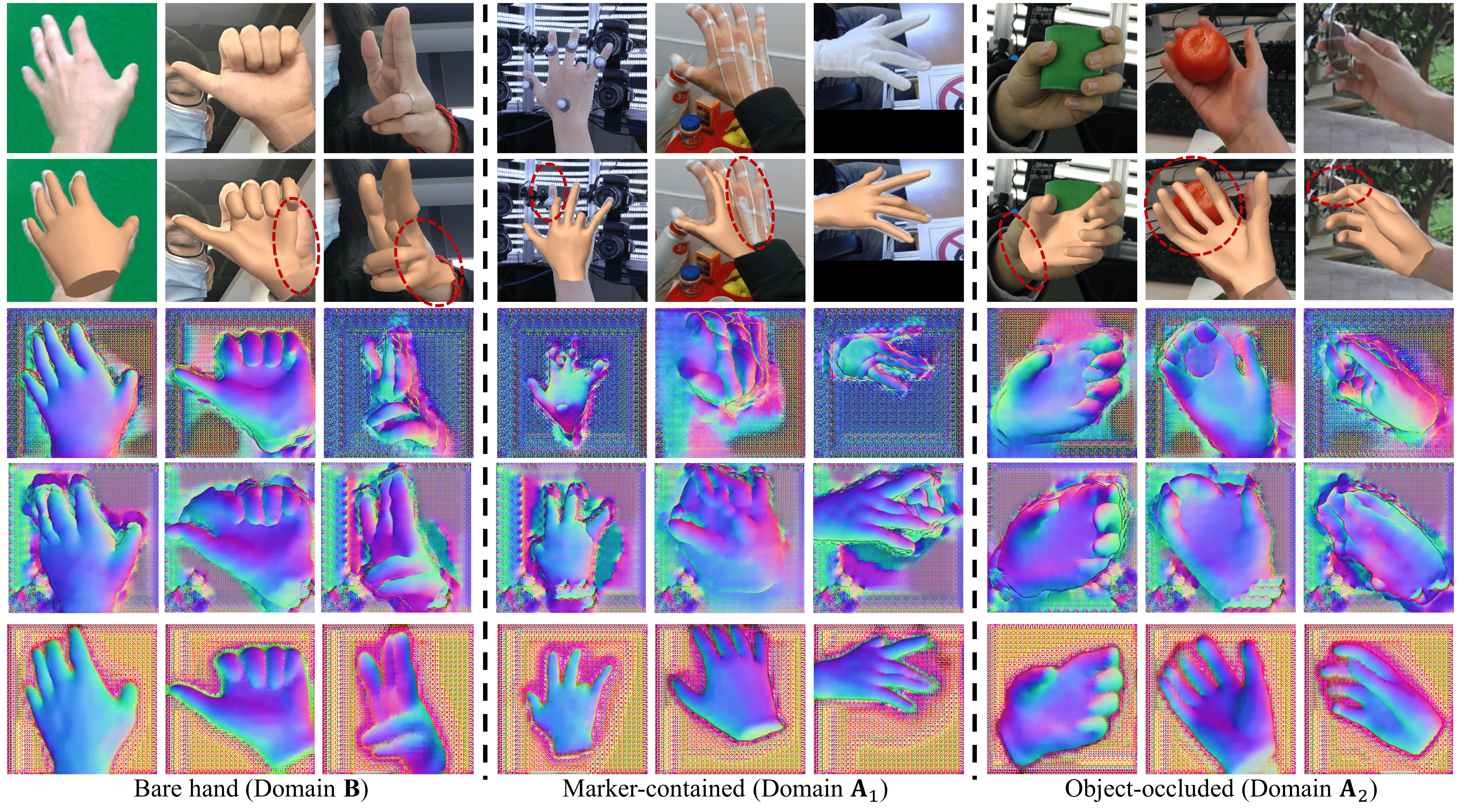}
    \vspace{-8mm}
    \caption{\textbf{Structure disentanglement from monocular RGBs}. (Row-1) Input images. (Row-2) Mesh recovery by a template-based strategy~\cite{zhou2020monocular}. (Row-3) Structure prediction by a template-free strategy~\cite{wang2018high}. (Row-4) Structure prediction by our sketcher w/o/ bare structure prior. (Row-5) Structure disentanglement by our full sketcher. Red circles indicate the artifacts in the results. }
    \vspace{-4mm}
    \label{fig08_compareNormal}
\end{figure*}

There are two strategies to wrap the appearance from one image to another. (i) Template-based strategies learn~\cite{siarohin2019animating, tang2018gesturegan, albahar2021pose} or optimize~\cite{munkberg2022extracting,pytorchthreed2017official} sophisticated wrappings based on parametric instance templates~\cite{romero2017embodied,qian2020html}. However, the accurate estimation of those parameters is heavily influenced by the degraded appearance in the images (See \figmk\ref{fig08_compareNormal} Row-2). (ii) Template-free ones~\cite{kolkin2019style, tumanyan2022splicing} excel at visible feature wrappings between structure-consistent images but are unable to selectively exclude marker-related features (See \figmk\ref{fig08_compareNormal} Row-3 and Row-4). To address the problem, we first embed the bare hand structure prior into pixel-aligned maps. Then this prior is encoded as the token form~\cite{bao2021beit}, and a ViT~\cite{dosovitskiy2020image} sketcher is trained to disentangle the corresponding structure tokens from partial image patches~\cite{he2022masked}. Interestingly, this ViT sketcher satisfies $S[S(X)] = S(X)$~\cite{fixedpoint1999wiki}, which means that when feeding its output as the input again, the two outputs should be consistent. We further utilize this elegant property to intensively train our sketcher in a semi-supervised paradigm. 

Disappointingly, the recovered appearances remain unsatisfactory when a structure-assisted translator trained with existing adversarial paradigms: (i) In popular supervised paradigms~\cite{isola2017image, wang2018high}, the discriminator focuses on the quality of the translation process. (ii) In most unsupervised paradigms~\cite{alami2018unsupervised,mueller2018ganerated, oprea2021h}, the discriminator can only evaluate the translation result since there is no reliable reference for the translation process. ~~Based on these two, we innovate the DAD scheme under a semi-supervised paradigm, which enables dual discrimination (both on the process and result) in our unpaired translation task. Initially, a partner domain is synthesized by degrading hand regions of the bare one. It possesses pairwise mapping relationships with the bare target domain, as well as similarity to the degraded source domain. During the translator training, data from the source and the partner domain are fed to the translator simultaneously. The two discriminators evaluate those translation processes and results with a clear division of labor. This scheme is more efficient than most unsupervised schemes~\cite{zhu2017unpaired, park2020contrastive} because of those trustworthy pairs. It is more generalizable than a supervised scheme trained only with synthetic degradation~\cite{li2018learning, li2020blind, wang2021towards} because of those multimodal inputs.

Our main contributions are summarized as follows.

\noindent$\bullet$ A semi-supervised framework that makes degraded images in marker-based \mocap regain bare appearance; 

\noindent$\bullet$ A powerful ViT sketcher that disentangles bare hand structure without parametric model dependencies;

\noindent$\bullet$ An adversarial scheme that promotes the degraded-to-bare appearance wrapping effectively.

\noindent The codes will be publicly available at ~\url{https://www.yangangwang.com}. 
\section{Related Work}
\noindent\textbf{Hand data capture}. 
Three procedures are widely used in hand capture: (i) Marker-based \mocap\cite{garcia2018first, han2018online, glauser2019interactive, sundaram2019learning, taheri2020grab, zhao2012combining} produces reliable motion but degraded appearance, so only the skeletal sequence is valuable for reconstruction. (ii) Synthetic appearance data~\cite{gao2022dart, hasson2019learning, lin2021two, yang2020seqhand, zimmermann2017learning} can be obtained from rendering digital hands~\cite{romero2017embodied,qian2020html}. However, the synthetic-to-real gap still exists even with the most advanced CG technology. (iii) Markerless \mocap\cite{zhang20163d, zhao2020hand, zimmermann2019freihand, moon2020interhand2} takes the goal to record motion without degrading appearance. It collects data in a multi-view stereo pipeline and performs learning-based pose estimation~\cite{wang2019srhandnet, lv2021handtailor, zhou2020monocular, chen2021camera} for each frame and each viewpoint. Although some weakly-supervised~\cite{spurr2021self, zhao2021travelnet, abdi20183d, cai2018weakly, kulon2020weakly, spurr2020weakly} paradigms are being explored, the dependency on dataset diversity and expensive annotation are still irreplaceable to train a robust estimator~\cite{armagan2020measuring}. Our framework recovers the marker-based \mocap data through image-to-image translation, which provides reliability for a estimator training.

\noindent\textbf{Human image synthesis}. 
The vision of a specific creature is highly discriminative of the authenticity of its own appearance. Thus, synthesizing photo-realistic human images is particularly challenging in both CV and CG communities. Differentiable rendering~\cite{munkberg2022extracting,pytorchthreed2017official} optimizes the whole render pipeline to minimize the difference between the result and the given images. However, in our single-view task, estimating either ambient lighting or human skin texture is difficult. Generative adversarial networks (GANs)~\cite{goodfellow2014generative} yield appealing performances to synthesize or recover human face~\cite{radford2015unsupervised, karras2019style, zhu2017unpaired, choi2018stargan}. This is primarily attributed to the fact~\cite{alami2018unsupervised, issenhuth2020not, albahar2021pose} that the facial images could be modeled as a 2D manifold with minor spatial deformation and easily aligned. When tackling other synthesis tasks involving highly nonlinear variations, most methods synthesize appearance based on object-specific templates, \forexample joint hierarchy~\cite{ma2017pose, tang2018gesturegan, siarohin2019animating, han2019clothflow}, part segmentations~\cite{guler2018densepose, grigorev2019coordinate, albahar2021pose} or surface topology~\cite{loper2015smpl}. Nevertheless, the presence of hand markers reduces the accuracy of estimating template parameters from images. We tackle this by embedding the bare hand structure prior in a pixel-aligned representation. 

\noindent\textbf{Unsupervised image-to-image translation}. 
Our task regards appearance recovery as a pixel-aligned image translation without pixel-aligned paired data. Some pioneers~\cite{mueller2018ganerated, oprea2021h} applied cycle consistency~\cite{zhu2017unpaired, kim2017learning, yi2017dualgan} to synthetic-to-real adaptation. However, the cycle paradigm models entire images without the semantics distinction and is not suitable for our task requiring partial modification. Some improvements~\cite{alami2018unsupervised, mueller2018ganerated, tang2021attentiongan} regard adaptive saliencies as attentions to assist the generator in focusing on partial semantics. Nonetheless, since neither supervision nor penalty is adopted for these saliencies, this may harm the generator. Recently, the attention expressed by dot product in latent space~\cite{vaswani2017attention,dosovitskiy2020image} has been widely adopted. Especially after a self-distillation pretraining~\cite{caron2021emerging}, these attentions~\cite{dosovitskiy2020image} show powerful semantics in correlations~\cite{amir2021deep} and appearance modeling~\cite{tumanyan2022splicing} without extra training or fine-tuning.
\begin{figure}[!t]
    \centering
    \includegraphics[width=\linewidth]{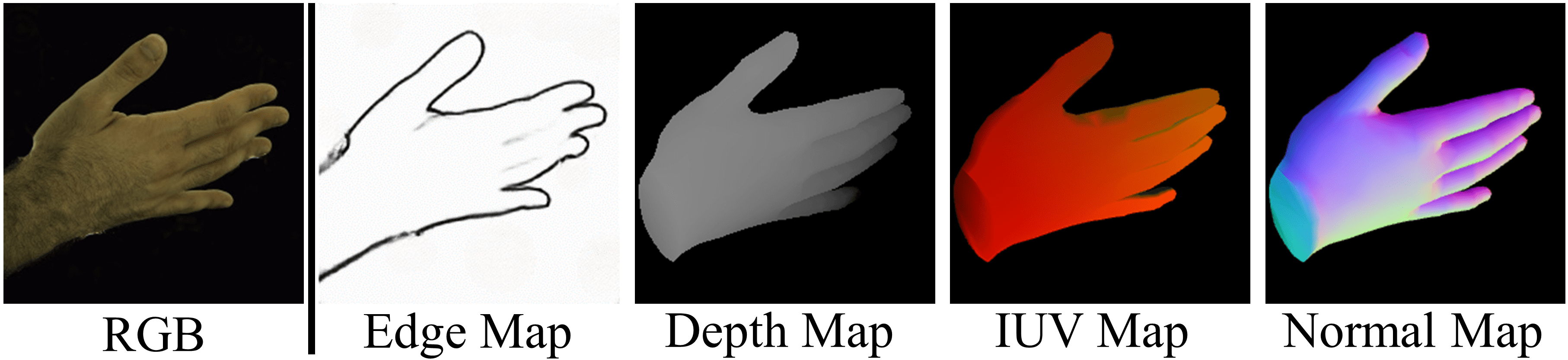}
    \vspace{-6mm}
    \caption{\textbf{Standardized domain candidates}. Edge map is estimated\cite{poma2020dense} from images. The others are rendered from meshes.}
    \vspace{-6mm}
    \label{fig02_structRep}
\end{figure}
\begin{figure*}[!t]
    \centering
    \includegraphics[width=\linewidth]{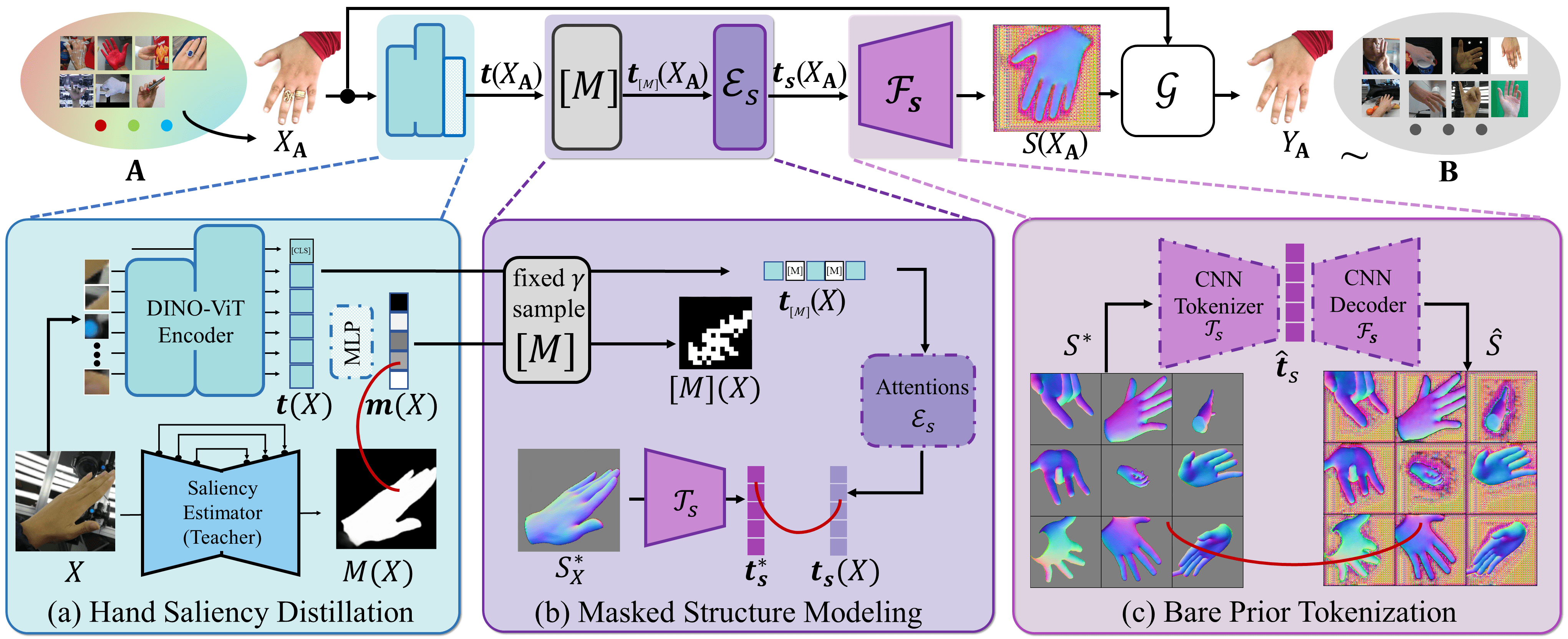}
    \caption{\textbf{Learning modules to structure disentanglement (better viewed in color)}. (a) \emph{Hand saliency distillation} gives the sketcher backbone extra attention for the hand. (b) \emph{Masked structure modeling} facilitates the sketcher's robustness to diverse degradations on hand structure. (c) \emph{Bare Prior Tokenization} makes the prior of the bare hand structure more compact in representation. In the training step of each module, the blocks in dashed are with learnable weights. The red curves illustrate the supervisions of training.}
    \vspace{-6mm}
    \label{fig04_modules}
\end{figure*}

\section{Method}
There exist two unpaired domains from a translation perspective: the source domain $\mathbf{A}$ refers to hand images $\{X_\mathbf{A}\}$ with diverse appearance degradations, and the target domain $\mathbf{B}$ refers to $\{X_\mathbf{B}\}$ with bare appearance. Synthetic data is not considered in our task. Two steps are taken to recover the hand appearance in $X_\mathbf{A} \in \mathbb{R}^{(3,h,w)}$: First, a sketcher disentangles the bare structure map $S(X_\mathbf{A})$ (\secmk\ref{sec31_structure}). After that, a translator $\mathcal{G}$ learns to wrap $S(X_\mathbf{A})$ with the appearance in $X_\mathbf{A}$ through our DAD scheme (\secmk\ref{sec32_semisupervised}). It is expected to generate $Y_\mathbf{A}$ by recovering the hand, reserving backgrounds, and removing degradations. 

\subsection{Structure Disentanglement}
\label{sec31_structure}
A desirable sketcher can disentangle the bare hand structure from images in both domains, so its input is no longer domain-distinct and generically denoted as $X$ in this section. As shown in \figmk\ref{fig04_modules}, the process is as follows: 
\begin{equation}
    \begin{aligned}
        X \rightarrow \bm{t}(X) \rightarrow \bm{t}_{[M]}(X) \rightarrow \bm{t}_{s}(X) \rightarrow S(X)
    \end{aligned}
    \label{eqn_demarker}
\end{equation}
\noindent\textbf{Saliency distillation}. $X \rightarrow\bm{t}(X)\in\mathbb{R}^{(n,d)}$ is executed in \dinovit\cite{caron2021emerging} by extracting the visual tokens from $n = (h/p)\cdot(w/p)$ non-overlapping image patches with uniform size $p\times p$. Because it is powerful to depict visible structure~\cite{amir2021deep, tumanyan2022splicing}, we adopt it as a frozen backbone. A four-layer MLP behind the backbone is adopted to regress patch-wise hand saliency $\bm{m}(X) \in [0,1]^{n}$. MLP is trained in a knowledge distillation scheme~\cite{hinton2015distilling,ge2021parser} by regarding a well-trained hand saliency estimator as the teacher, who estimates $M(X) \in [0,1]^{(h,w)}$ (See \supmat for details). 

\noindent\textbf{Structure domain}. $S(X) \in \mathbb{R}^{(c,h,w)}$ is defined under a standardized domain. \figmk\ref{fig02_structRep} enumerates some candidates. The edge map is first excluded because it is too sparse. The other three can be acquired by rendering hand models~\cite{romero2017embodied, moon2020deephandmesh,zhao2021supple}, which frees us from the dependence on real images to construct the bare structure prior. Among them, values in a depth map may be ambiguous in our single-view task. The creation of an IUV map additionally depends on the fixed UV unwrapping. By comparison, a normal map ($c$\Equal3) circumvents most shortcomings. Based on it, we render a dataset $\mathbf{S}$ containing \datanum diverse instances. 

\noindent\textbf{Prior tokenization}. The selected domain that lacks an instance-specific prior is still too redundant. To make it more compact to represent a bare hand and also save the modeling capability of ViT, a discrete VAE~\cite{bao2021beit, ramesh2021zero} $\{\mathcal{T}_{s}, \mathcal{F}_{s}\}$ is introduced to enable a map-token conversion $S\leftrightarrows \bm{t}_s$. It is trained with the data $S^\star \in \mathbf{S}$ by the following loss:
\begin{equation}
    \begin{aligned}
        L_{\mathcal{T}_{s}, \mathcal{F}_{s}} = \lambda_{k} L_{\mathrm{kl}} (\bm{\hat{t}}_s) + \|\hat{S} - S^\star\|_F + \lambda_{g} L_{\mathrm{ad}}[\hat{S} , S^\star] 
    \end{aligned}
    \label{eqn_tokenloss}
\end{equation}
where $\|\cdot\|_F$ is the Frobenius norm. The encoded tokens $\bm{\hat{t}}_s$\Equal$\mathcal{T}_{s}(S^\star)$ are regulated as uniform distribution. The mean squared error (MSE) between the decoded map $\hat{S}$\Equal$\mathcal{F}_{s} (\bm{\hat{t}}_s)$ and $S^\star$ preserves the low-frequency details, while the adversarial term $ L_{\mathrm{ad}}$ borrowed from~\cite{isola2017image} preserves its high-frequency details. 

\noindent\textbf{Masked modeling}. Supervised by the well-trained tokenizer $\mathcal{T}_{s}$, a na\"ive baseline is to perform $\bm{t}(X) \rightarrow \bm{t}_{s}(X)$ with annotated data $\{X, S^\star_X\}$ in degradation-contained datasets~\cite{zimmermann2019freihand, hampali2020honnotate, zhao2020hand}. However, due to their limited amount, such a process may harm the model's generalization. We further recast it as masked image modeling (MIM) and introduce a mask-guided learning strategy. Instead of a random formulation~\cite{bao2021beit, he2022masked}, we sample a fixed ratio $\gamma$ of the tokens $\left[M\right](X)$ to be masked out according to a multinomial distribution related to the patch hand saliency $\bm{m}(X)$: 
\begin{equation}
    \begin{aligned}
        \left[M\right](X) \sim \mathrm{multinomial} [1 - \bm{m}(X)+ \varepsilon ; \gamma]
    \end{aligned}
    \label{eqn_smapling}
\end{equation}
where $\varepsilon$\Equal$10^{-5}$ confirms the nor-zero probability of all patches. As a result, $\bm{t}(X) \rightarrow \bm{t}_{[M]}(X) \in \mathbb{R}^{(n,d)}$ is realized by replacing $(\gamma\cdot n)$ samples as the same learnable mask token. 

After that, a ViT decoder $\mathcal{E}_s$ learns the conversion of $\bm{t}_{[M]}(X) \rightarrow \bm{t}_{s}(X)$. We first train it with $\{X, S^\star_X\}$ according to the following loss: 
\begin{equation}
    \begin{aligned}
        L_{\mathcal{E}_s} &= \|\mathcal{E}_{s}(\bm{t}_{[M]}(X)) - \bm{t}_s^\star \|_F +
        \|\mathcal{E}_{s}(\bm{t}_{[M]}(S^\star_X)) - \bm{t}_s^\star \|_F \\
    \end{aligned}
    \label{eqn_vitStruct}
\end{equation}
where $\bm{t}_s^\star = \mathcal{T}_{s}(S^\star_X)$. The second term is introduced because the standardized domain is a special image domain containing bare hand structure. This means that an expected structure map $\overline{S(X)}$ extracted from an image $X$ should always be a fixed point~\cite{fixedpoint1999wiki} for our sketcher: 
\begin{equation}
    \begin{aligned}
        S[\overline{S(X)}] = \overline{S(X)}
    \end{aligned}
    \label{eqn_semiS}
\end{equation}
Based on \equationmk\ref{eqn_semiS}, our sketcher is further fine-tuned with those datasets $\{X\}$ without available $S^\star_X$ as follows: 
\begin{equation}
    \begin{aligned}
        L_{\mathcal{E}_s, \mathcal{F}_s} = \| S(X) - S[S(X)] \|_F
    \end{aligned}
    \label{eqn_ef2}
\end{equation}
This semi-supervised paradigm significantly strengthens the collaboration between $\mathcal{E}_s$ and $\mathcal{F}_s$, which are always trained separately in the existing literature~\cite{bao2021beit, ramesh2021zero}. To sum up, the whole sketcher disentangles the hand structure map by the following produce: 
\begin{equation}
    \begin{aligned}
        S(X) = \mathcal{F}_{s} \big( \mathcal{E}_{s}[\bm{t}_{[M]}(X)]\big)
    \end{aligned}
    \label{eqn_sketcher}
\end{equation}

\noindent\textbf{Implementation details}. All pieces of training are optimized by Adam at a base learning rate of $10^{-4}$ and a batch size of $16$. We use \dinovit with $p$\Equal$16$ as the frozen backbone. We set images with size $h$\Equal$w$\Equal$256$, the codebook with size $512$, and $\bm{t}_s \in \mathbb{R}^{(512,16,16)}$. The backbone of $\mathcal{T}_{s}$ is ResNet50, and $\mathcal{F}$ is built symmetrically by transpose convolutions. $L_{ad}$ is computed discriminatively by a multi-scale patchGAN~\cite{isola2017image}. The adversarial weight $\lambda_{g}$ is set at $0.01$ constantly. The KL weight $ \lambda_{k}$ is increased from 0 to $6.6 \times 10^{-7}$ and Gumbel-SoftMax~\cite{jang2016categorical, maddison2016concrete} relaxes temperature $\tau$ from 1.0 to 0.5 over the first 5000 updates. ~~$\mathcal{E}_{s}$ is built as $12$ attention blocks with fixed sin-cos position embeddings~\cite{vaswani2017attention}. The masking-out ratio is set to $\gamma$\Equal$0.75$. 

\begin{figure}[!t]
    \centering
    \includegraphics[width=\linewidth]{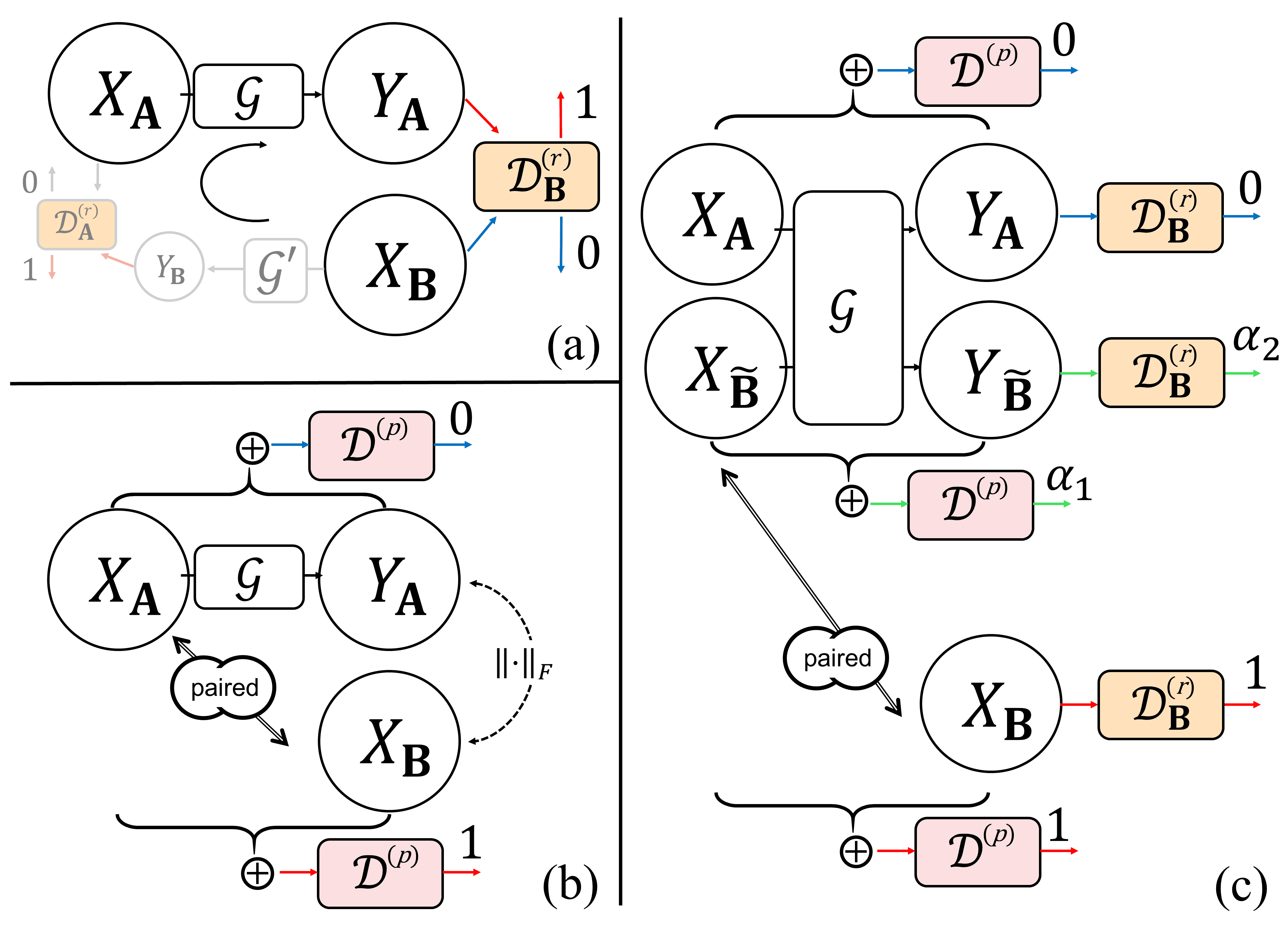}
    \vspace{-6mm}
    \caption{\textbf{Learning paradigm comparisons.} (a) CycleGAN~\cite{zhu2017unpaired} in an unsupervised paradigm. (b) Pix2pix~\cite{isola2017image} in a supervised paradigm. (c) Our DAD in a semi-supervised paradigm. In each sub-figure, blocks of the same color share weights.}
    \vspace{-6mm}
    \label{fig03_paradigm}
\end{figure}

\subsection{Appearance Wrapping}
\label{sec32_semisupervised}
The sketcher disentangles the bare structure $S(X_\mathbf{A})$ containing both visible and degraded hand parts. Next, a wrapper learns to map valid appearance from $X_\mathbf{A}$ to $S(X_\mathbf{A})$, achieving hand appearance recovery on its output $Y_\mathbf{A}$. 

\noindent\textbf{Paradigm evolutions}. Before introducing our dual adversarial discrimination (DAD) in a semi-supervised paradigm, we progressively add related components. (i) The unsupervised paradigm~\cite{zhu2017unpaired} is shown in \figmk\ref{fig03_paradigm}(a). Its adversarial scheme introduces \emph{result discriminator}(s) $\mathcal{D}_\mathbf{B}^{(r)}, \mathcal{D}_\mathbf{A}^{(r)}$, which enables $\mathcal{G}$ and the inverse translator $\mathcal{G}^\prime$ to form a bijection$\mathbf{A}\leftrightarrows\mathbf{B}$. (ii) The supervised one~\cite{isola2017image} is shown in \figmk\ref{fig03_paradigm}(b). Its adversarial scheme has a \emph{process discriminator} $\mathcal{D}^{(p)}$, which prompts $\mathcal{G}$ to learn more details based on computable MSE among paired data. Inspired by face restorations~\cite{li2018learning, li2020blind, wang2021towards}, we synthesize a partner domain $\tilde{\mathbf{B}}$ by degrading $X_\mathbf{B} \in \mathbf{B}$ with diverse noise $X_{\tilde{\mathbf{B}}}$\Equal$\tilde{N}(X_\mathbf{B})$ (See \supmat for details). This paired degradation process $\mathbf{B} \rightarrow \tilde{\mathbf{B}} $ makes this supervised paradigm feasible.

\noindent\textbf{DAD scheme}. Our semi-supervised paradigm shown in \figmk\ref{fig03_paradigm}(c) introduces $\mathcal{D}_\mathbf{B}^{(r)}, \mathcal{D}^{(p)}$ together to judge a multimodal translation $(\tilde{\mathbf{B}}, \mathbf{A}) \rightarrow \mathbf{B}$ on both result qualities and process qualities. We also utilize the above-mentioned partner $X_{\tilde{\mathbf{B}}}$\Equal$\tilde{N}(X_\mathbf{B})$ and optimize $\mathcal{G}$ through the following loss: 
\begin{equation}
    \begin{aligned}
        L_{\mathcal{G}} &= \|(Y_\mathbf{A} \shortminus X_A) \odot (1 \shortminus M[S(X_\mathbf{A})])\|_F \\
        &+\|(Y_{\tilde{\mathbf{B}}} \shortminus X_{\tilde{\mathbf{B}}})\odot (1 \shortminus M[S(X_{\tilde{\mathbf{B}}})]) \|_F \\
        &+|\mathcal{D}_\mathbf{B}^{(r)}(Y_\mathbf{A})\shortminus 1| + |\mathcal{D}_\mathbf{B}^{(r)}(Y_{\tilde{\mathbf{B}}})\shortminus 1|\\
        &+|\mathcal{D}^{(p)}(X_\mathbf{A} \oplus Y_\mathbf{A})\shortminus 1| + |\mathcal{D}^{(p)}(X_\mathbf{\tilde{\mathbf{B}}} \oplus Y_{\tilde{\mathbf{B}}})\shortminus 1|\\
    \end{aligned}
    \label{eqn_lossG}
\end{equation}
where $Y_\mathbf{A}$\Equal$\mathcal{G}(X_\mathbf{A}), Y_{\tilde{\mathbf{B}}}$\Equal$\mathcal{G}(X_{\tilde{\mathbf{B}}})$, $\oplus$ is channel-wise concatenation, $\odot$ is element-wise product. The first two terms preserve the non-hand semantics outside $M[S(X)]$. The next two encourage $\mathcal{G}$ to fool $\mathcal{D}_\mathbf{B}^{(r)}$, and the last two to fool $\mathcal{D}^{(p)}$. Adversarially, the two discriminators are trained by: 
\begin{equation}
    \left\{
    \begin{aligned}
        L_{\mathcal{D}}^{(r)} &= |\mathcal{D}_\mathbf{B}^{(r)}(\underline{Y_\mathbf{A}})\shortminus 0| + |\mathcal{D}_\mathbf{B}^{(r)}(\underline{Y_{\tilde{\mathbf{B}}}})\shortminus \alpha_2| + |\mathcal{D}_\mathbf{B}^{(r)}(X_\mathbf{B})\shortminus 1| \\
        L_{\mathcal{D}}^{(p)} &= |\mathcal{D}^{(p)}(X_\mathbf{A} \oplus \underline{Y_\mathbf{A}})\shortminus 0| + |\mathcal{D}^{(p)}(X_{\tilde{\mathbf{B}}} \oplus \underline{Y_{\tilde{\mathbf{B}}}})\shortminus \alpha_1|\\
        &+ |\mathcal{D}^{(p)}(X_{\tilde{\mathbf{B}}} \oplus X_\mathbf{B})\shortminus 1|\\
    \end{aligned}
    \right.
    \label{eqn_losstD}
\end{equation}
where $\underline{Y}$ denotes stop-gradient. Two tolerances $\alpha_1, \alpha_2$ sampled uniformly from $U(0.4, 0.7)$ are the scores of those plausible but synthesized translations. 

\begin{figure}[!t]
    \centering
    \includegraphics[width=\linewidth]{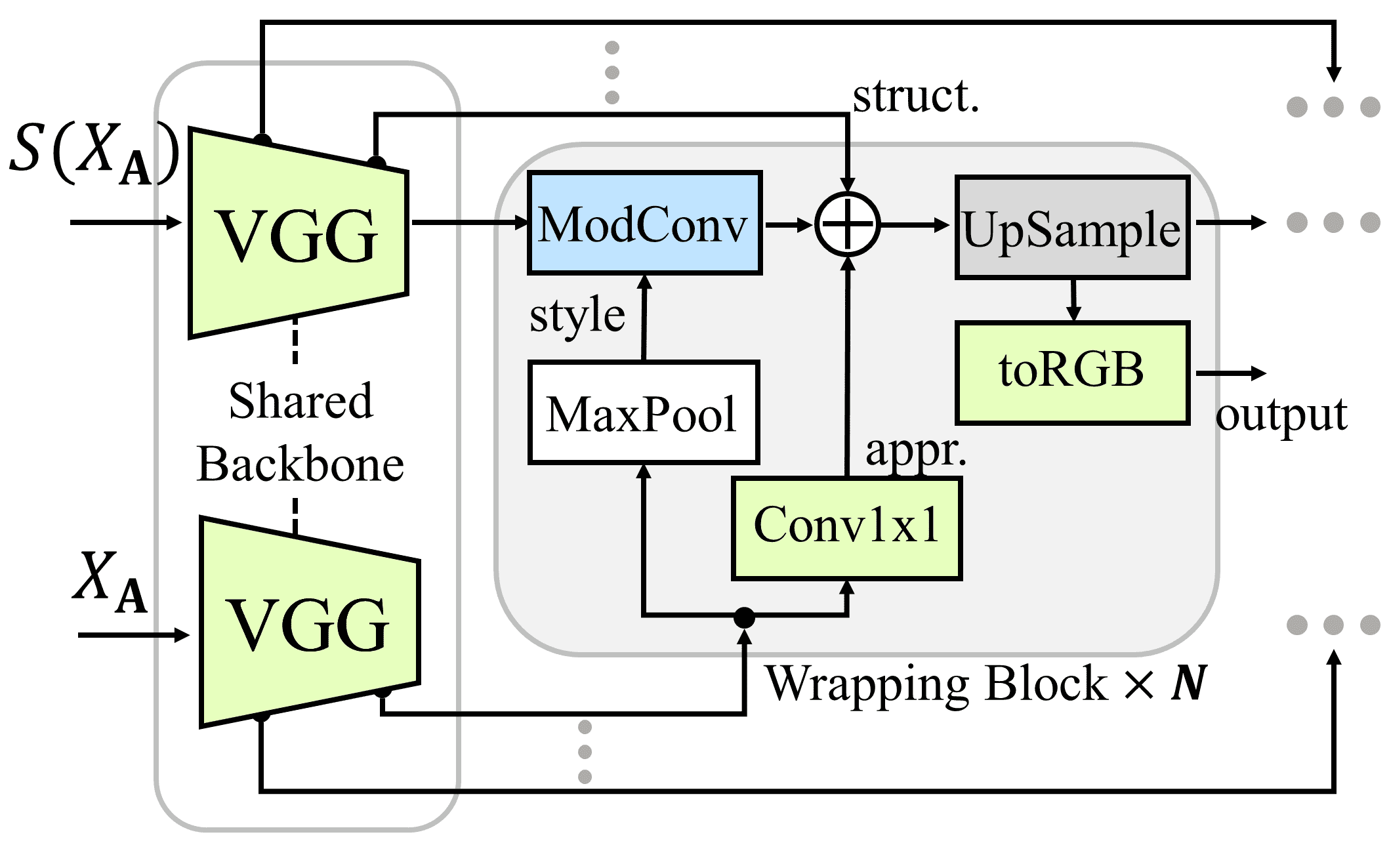}
    \vspace{-6mm}
    \caption{\textbf{Translator architecture}. \mybox{ModConv} is the modulated convolution~\cite{karras2020analyzing}. \mybox{Conv1x1} is the convolution with a kernel size of 1. \mybox{toRGB} is the convolution that converts the multi-channel feature to the three-channel feature with a kernel size of 1.}
    \vspace{-6mm}
    \label{fig11_translatorArch}
\end{figure}

\noindent\textbf{Translator Architecture}. As illustrated in \figmk\ref{fig11_translatorArch}, our translator takes $X_\mathbf{A}, S(X_\mathbf{A})$ as the inputs and separately extracts their multi-level features with a shared CNN backbone. After that, $\mathcal{G}$ gradually fuses the structure and appearance details at the same level with $N$ wrappers based on the deepest features of $S(X_\mathbf{A})$. The internal structure of the wrapper inherits from the synthesis layer in StyleGAN2~\cite{karras2020analyzing}. In each level, a max-pooling of the appearance feature provides the style to the modulated convolution, and a 1x1 convolution layer learns to filter the appearance and discard the degradation. The image mapping (``toRGB'') is added at each level, which makes the training more efficiently~\cite{karras2020analyzing}. And the mapping at the final level outputs $Y_\mathbf{A}$. 

\noindent\textbf{Implementation details}.  The training is optimized by Adam at a base learning rate of $10^{-4}$ and a batch size of $16$. The convolution backbone is selected as VGG-16~\cite{simonyan2014very}. The image mapping at each level is supervised by the first two terms in \equationmk\ref{eqn_lossG} in their resolutions. We inherit~\cite{kolkin2019style} to take VGG features in $[1,3,5,10,13]$ layers. Consequently, $N$\Equal$5$ wrappers are adopted in the following process. 

\begin{figure}[!t]
    \centering
    \includegraphics[width=\linewidth]{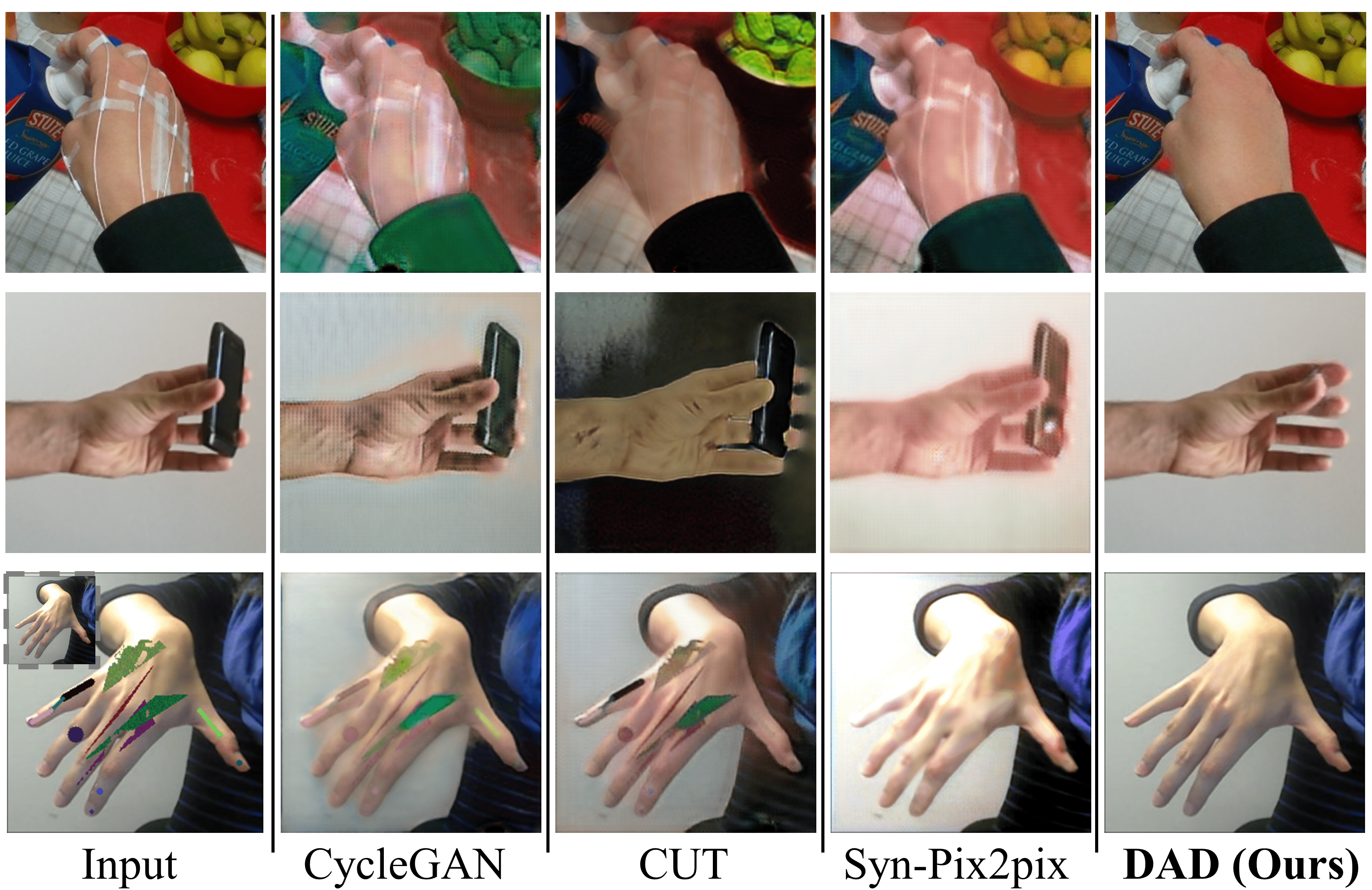}
    \caption{\textbf{Comparisons on learning paradigms}. From left to right: The input, CycleGAN~\cite{zhu2017unpaired}, CUT~\cite{park2020contrastive}, Syn-Pix2pix~\cite{wang2018high}, and Ours. The testing samples are from domain $\mathbf{A}_{\text{1}}$, $\mathbf{A}_{\text{2}}$ and $\tilde{\mathbf{B}}$. The original version of the sample from $\tilde{\mathbf{B}}$ is on its top left.
    }
    \vspace{-4mm}
    \label{fig07_compareCyclegan}
\end{figure}
\section{Experiments}
The baselines related to our appearance recovery task are enumerated in \secmk\ref{sec41_baseline}. The adopted data and metrics for training and evaluations are then presented in \secmk\ref{sec42_data}. Different frameworks are compared in \secmk\ref{sec43_comp}, and our key components are ablated in \secmk\ref{sec44_ablation}. 

\subsection{Baselines}
\label{sec41_baseline}
\noindent\textbf{Image translation} is used to formulate our hand appearance recovery task. We make comparisons to (i) CycleGAN~\cite{zhu2017unpaired} and its successors, including GANerated~\cite{mueller2018ganerated}, UAG~\cite{alami2018unsupervised} and H-GAN~\cite{oprea2021h}. (ii) CUT~\cite{park2020contrastive} models the problem with contrastive strategies. (iii) Since an additional partner domain data $\tilde{\mathbf{B}}$ is introduced in the DAD scheme, we also compare the translating performance trained in a full supervised paradigm~\cite{wang2018high} with paired data $\tilde{\mathbf{B}}\leftrightarrows \mathbf{B}$. It is denoted as "Syn-Pix2pix". 

\noindent\textbf{Differentiable rendering} optimizes parametric model~\cite{romero2017embodied}, texture~\cite{qian2020html} and lighting in a rendering pipeline to minimize the visual differences between output and given image~\cite{pytorchthreed2017official}. To make it executable, we manually make extra annotations in its testing: (i) left or right side, (ii) 2D key points for pose optimization, and (iii) silhouette for shape optimization. This pipeline is denoted as "Diff-Render". 

\noindent\textbf{Neural style transfer (NST)} takes appearance and structure images as separated inputs~\cite{gatys2016image}. For a fair comparison, we overlay $S(X)$ to a testing image $X$ as their structure reference ($X$ provides extra background structure), and use the original $X$ as their appearance reference. Three representative works are selected: WCT2~\cite{yoo2019photorealistic} is highly effective. STROTSS~\cite{kolkin2019style} utilizes VGG features as feedback. \splicevit\cite{tumanyan2022splicing} utilizes ViT features as feedback. 

\begin{table*}[!t]
    \rowcolors{1}{}{lightgray}
    \begin{center}
        \resizebox{1\linewidth}{!}{
    \begin{tabular}{c|cc|cc|cc|cc}
    \noalign{\hrule height 1.5pt}
    Tasks 
    &\multicolumn{4}{c|}{$\mathbf{A}_{\text{1}} \rightarrow \mathbf{B}$}
    &\multicolumn{4}{c}{$\mathbf{A}_{\text{2}} \rightarrow \mathbf{B}$}\\
    \midrule
    \rowcolor{white}Metrics 
    &$\text{FID}_{\text{i}} \downarrow$ &$\text{KID}_{\text{i}}(*100) \downarrow$ &$\text{FID}_{\text{v}}\downarrow$ &$\text{KID}_{\text{v}}\downarrow$
    &$\text{FID}_{\text{i}}\downarrow$ &$\text{KID}_{\text{i}}(*100)\downarrow$ &$\text{FID}_{\text{v}}\downarrow$ &$\text{KID}_{\text{v}}\downarrow$ \\ 
    \midrule
    \midrule
    CycleGAN~\cite{zhu2017unpaired}   
    &$76.39$     &$4.46\pm0.176$    &$1266.17$   &$32.02\pm0.994$
    &$65.12$     &$4.44\pm0.196$    &$1021.14$    &$27.13\pm0.948$\\
    GANerated~\cite{mueller2018ganerated}
    &$76.97$     &$4.72\pm0.153$    &$1220.53$   &$31.72\pm0.907$ 
    &$68.50$     &$5.05\pm0.199$    &$985.82$    &$26.75\pm0.987$\\
    H-GAN~\cite{oprea2021h} 
    &$93.02$    &$6.53\pm0.209$     &$1488.32$   &$37.85\pm1.061$ 
    &$62.94$     &$4.12\pm0.197$    &$876.94$    &$24.61\pm0.816$\\
    UAG~\cite{alami2018unsupervised}  
    &$87.80$    &$5.90\pm0.177$     &$1375.55$   &$35.67\pm0.949$ 
    &$70.98$     &$5.35\pm0.210$    &$1069.45$   &$28.31\pm0.926$\\
    CUT~\cite{park2020contrastive}
    &$78.02$     &$5.54\pm0.192$    &$1230.67$   &$33.34\pm1.015$
    &$58.88$     &$3.73\pm0.160$    &$749.22$   &$20.02\pm0.826$\\
    Ours
    &$\bm{60.37}$     &$\bm{3.45\pm0.236}$    &$\bm{994.67}$   &$\bm{28.67\pm0.916}$
    &$\bm{41.53}$     &$\bm{3.37\pm0.154}$    &$\bm{673.43}$   &$\bm{15.72\pm1.209}$ \\
    \noalign{\hrule height 1.5pt}
    \end{tabular}
    }
    \end{center}
    \vspace{-4mm}
    \caption{\textbf{Evaluations for Translation.} The first column describes the translation framework. The next two column groups describe the performance of each framework on $\mathbf{A}_{\text{1}} \rightarrow \mathbf{B}$ and $\mathbf{A}_{\text{2}} \rightarrow \mathbf{B}$ respectively. }
    \label{tab01_compareTranslation}
    \vspace{-4mm}
\end{table*}

\subsection{Datasets and Metrics}
\label{sec42_data}

\begin{figure}[!t]
    \centering
    \includegraphics[width=\linewidth]{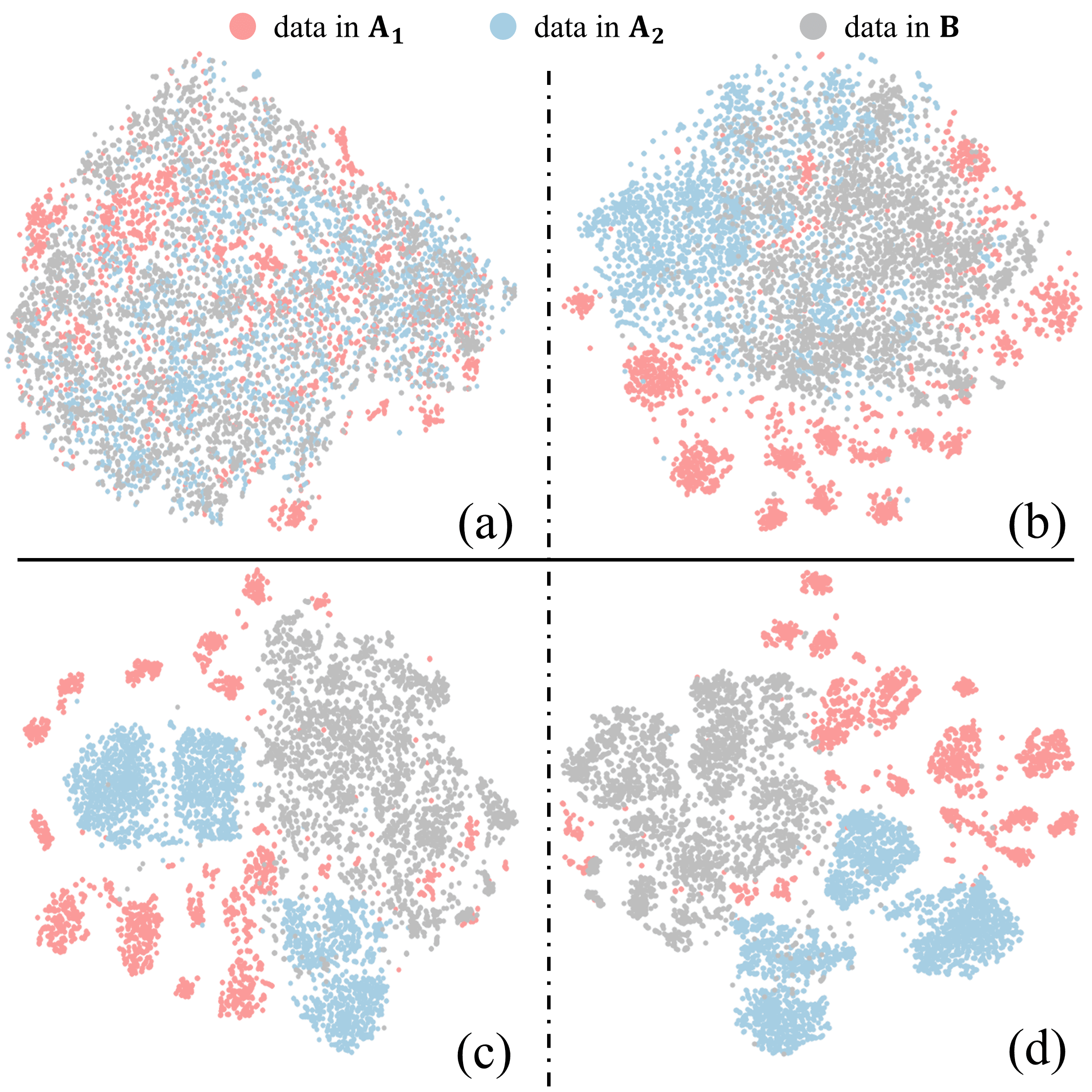}
    \caption{\textbf{Feature distributions of domain datasets (better viewed in color)}. (a) InceptionV3 pre-aux features(dim=$768$). (b) InceptionV3 final features (dim=$2048$). (c) DINO-ViT-b16 class tokens (dim=$768$). (d) DINO-ViT-b8 class tokens (dim=$768$). }
    \vspace{-4mm}
    \label{fig06_tsne}
\end{figure}

\noindent\textbf{Structure data}. To tokenize bare structure prior with $\{\mathcal{T}_{s}, \mathcal{F}_{s}\}$, depth, IUV, or normal maps are obtained through graphic rendering. Depth map~$S_d$ and normal map~$S_n$ only need mesh data. IUV map~$S_{uv}$ uses the UV unwrapping in~\cite{qian2020html}. \datanum samples for each domain are created using hand parametric models~\cite{romero2017embodied}, pose archive~\cite{yang2020seqhand, zhao2021travelnet} and publicly available scans~\cite{moon2020deephandmesh}. The data used to train $\mathcal{E}_{s}$ contains both annotated datasets $\{X, S^\star_X\}$ (Freihand~\cite{zimmermann2019freihand}, HandStudio~\cite{zhao2020hand}, HO3D~\cite{hampali2020honnotate}) and unlabeled ones $\{X\}$(Core50~\cite{lomonaco2017core50}, EgoDexter~\cite{mueller2017real}, DexterObject~\cite{sridhar2016real}).

\noindent\textbf{Appearance data}. Each of the following domains are represented by $33$K image data ($30$K for training, $3$K for evaluation). They are sampled from diverse datasets, and contain a comparable number of images in complex or monochrome backgrounds. All images are cropped to be hand-centered. 

\noindent$\bullet$ Marker-contained domain ($\mathbf{A}_{\text{1}}$) covers a wide range of adhesion markers: FPHAB~\cite{garcia2018first} (inertia), MHP~\cite{gomez2019large} (gloves), PaintedHand~\cite{mueller2019real} (paintings), and 11kHands~\cite{afifi201911k} (rings). We further collect data containing optical markers~\cite{han2018online}.

\noindent$\bullet$ Object-occluded domain ($\mathbf{A}_{\text{2}}$) refers to the images in which the object occludes the hand, rather than any hand-object interacting images. They are sifted manually from Freihand~\cite{zimmermann2019freihand}, HandStudio~\cite{zhao2020hand}, HO3D~\cite{hampali2020honnotate}, Core50~\cite{lomonaco2017core50}, EgoDexter~\cite{mueller2017real} and DexterObject~\cite{sridhar2016real}. 

\noindent$\bullet$ Bare hand domain ($\mathbf{B}$) data is gathered from OneHand10k~\cite{wang2018mask}, DMTL~\cite{zhang2021hand}, HandStudio~\cite{zhao2020hand}, STB~\cite{zhang2017hand}, InterHand2.6M~\cite{moon2020interhand2} and GANerated~\cite{mueller2018ganerated}.

\noindent\textbf{Human perceptual metrics.} The human perceptual survey is performed on Amazon Mechanical Turk (AMT): Each question contains degraded input and the recovering results from random frameworks in a row. Each Turker is assigned 30-40 questions. This survey is performed with 90 images gathered from the testing samples in degraded domains. They are reported in our \supmat. 

\noindent\textbf{DNN perceptual metrics}. Frechet Inception Distance~\cite{heusel2017gans} (FID) and Kernel Inception Distance~\cite{binkowski2018demystifying} (KID) measure the difference between translated results and target domain. Because they evaluate images in a latent space, an image feature extractor is required. Considering that CNN and ViT have a different emphasis, both InceptionV3~\cite{szegedy2016rethinking} and \dinovit are adopted. We explore their distinguishability on our data domains through t-SNE~\cite{hinton2002stochastic}. As shown in \figmk\ref{fig06_tsne}, InceptionV3 final features and DINO-ViT-b8 class tokens describe the cross-domain relationships more clearly. Consequently, FID and KID based on the two features are denoted as $(\text{FID}_{\text{i}}, \text{KID}_{\text{i}})$ and $(\text{FID}_{\text{v}}, \text{KID}_{\text{v}})$, respectively. This choice also takes into account that our sketcher's backbone is DINO-ViT-b16 (base architecture, patch size 16), so the evaluation with DINO-ViT-b8 leads to a fair comparison. In addition, all image backgrounds are filtered with $M[S(X)]$ before DNN feature extraction. 

\begin{figure}[!t]
    \centering
    \includegraphics[width=\linewidth]{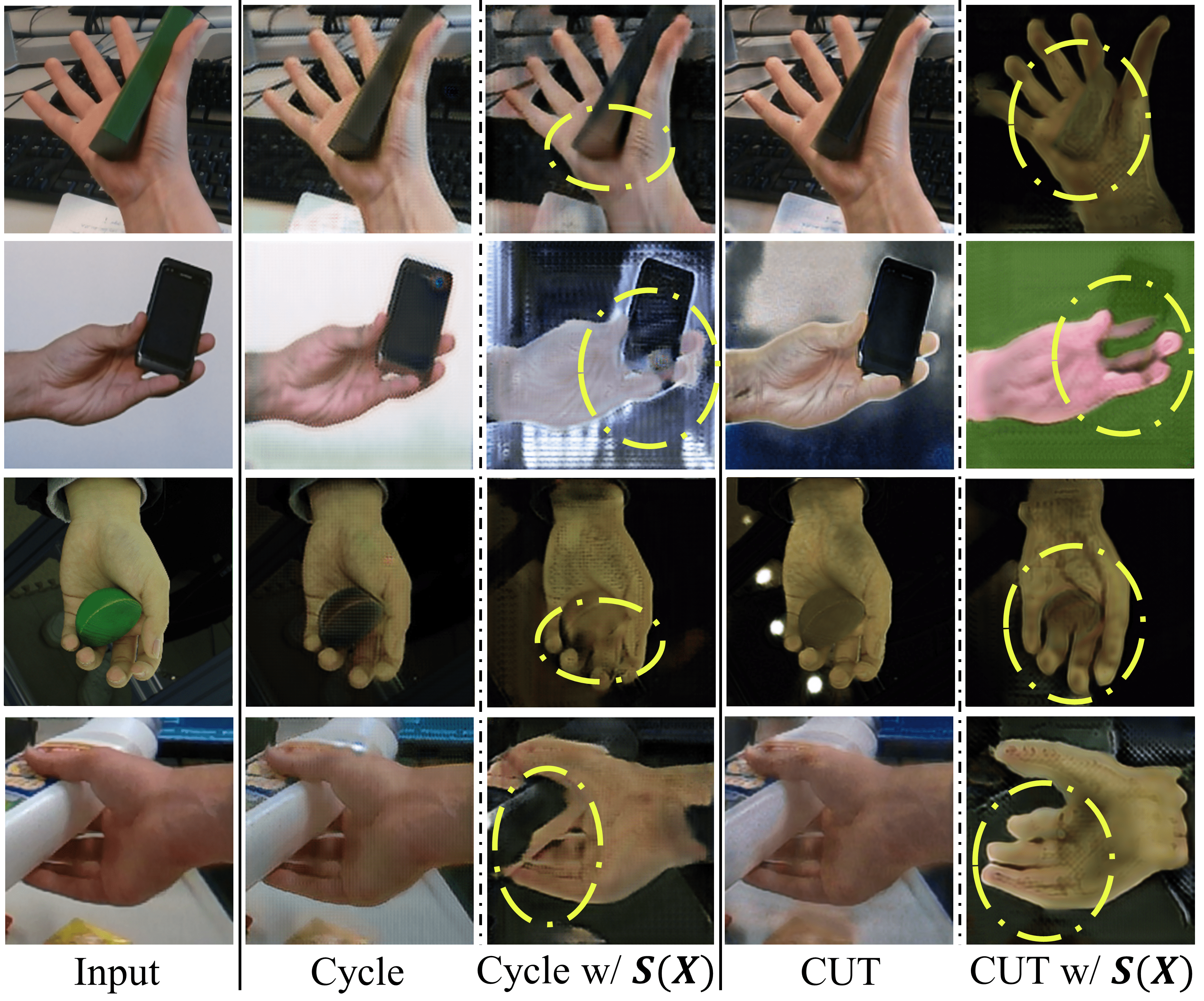}
    \caption{\textbf{Influence of the disentangled structure}. From left to right: The input, CycleGAN~\cite{zhu2017unpaired} w/ or w/o/ $S(X)$ assistance, and CUT~\cite{zhu2017unpaired} w/ or w/o/ $S(X)$ assistance. The yellow circles indicate the improvements in each group brought by our bare structure. }
    \vspace{-4mm}
    \label{fig04_ablationSX}
\end{figure}

\subsection{Comparisons}
\label{sec43_comp}

\begin{figure}[!t]
    \centering
    \includegraphics[width=\linewidth]{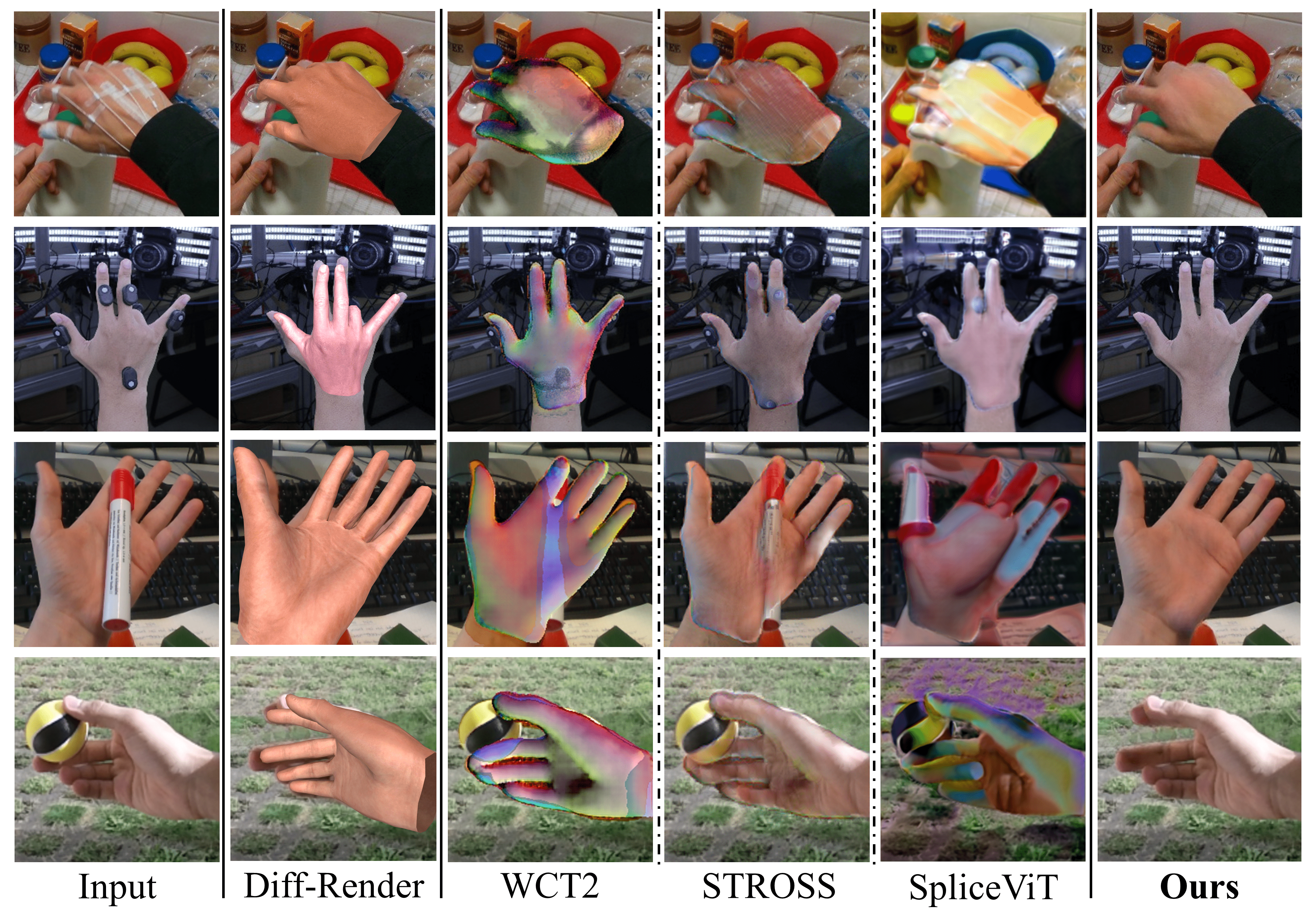}
    \caption{\textbf{Comparisons to the SOTA NSTs}. From left to right: The input, Diff-Render, WCT2~\cite{yoo2019photorealistic}, STROTSS~\cite{kolkin2019style}, \splicevit\cite{tumanyan2022splicing}, and Ours. The first two rows are samples from $\mathbf{A}_{\text{1}}$ and the last two rows from $\mathbf{A}_{\text{2}}$.}
    \vspace{-4mm}
    \label{fig07_compareNST}
\end{figure}

\noindent\textbf{Qualitative comparisons}. We compare with existing unsupervised frameworks in three aspects: (i) Full pipeline: The results in \figmk\ref{teaser_optRes} are from their original paradigm+translator. (ii) Paradigm only: The results in \figmk\ref{fig07_compareCyclegan} are from their paradigms+our $\mathcal{G}$. (iii) Structure influence: The results in \figmk\ref{fig04_ablationSX} are from their full pipeline w/ or w/o our $S(X)$ as an additional input. ~~In (i) and (ii), they neither guarantee the hand bareness nor preserve the background during translation. In (iii), although their results are still not appealing enough, they do have a significant improvement in hand structure. ~~For the sub-task of structure disentanglement, our prior-based structure sketcher is compared with a template-based pose estimator~\cite{zhou2020monocular} and a template-free image translator~\cite{wang2018high}. The estimator needs to convert all hands in the image to left-handed as preprocessing. The translator relies on accurate hand saliencies to filter out non-hand features. More conveniently, our sketcher takes the image directly as input, and the internal ViT samples the patch adaptively. As shown in \figmk\ref{fig08_compareNormal}, our sketcher outperforms the pose estimator in hand part localization. It also ensures hand bareness to a greater extent while being robust to backgrounds and degradations. ~~For the sub-task of appearance wrapping, we compare with Diff-Render and NSTs by feeding $X$ and $S(X)$ to them. \figmk\ref{fig07_compareNST} shows their performances on several testing samples. The synthesis-to-real gap is inevitable in Diff-Render results. Some skin tones and textures are aliased in NSTs. 

\noindent\textbf{Quantitative comparisons}. \tablemk\ref{tab01_compareTranslation} extensively illustrates the recovery qualities of the unsupervised frameworks and ours. FIDs and KIDs quantitatively evaluate results on the differences and variances to domain $\mathbf{B}$ in the latent space. We outperform other methods under both ViT- and CNN- based criteria, and the gap is more obvious in the criteria described by ViT. This reveals the rationality of modeling the degraded hand appearance with ViT. From \tablemk\ref{tab01_compareTranslation}, the performance of those mask-based methods~\cite{mueller2018ganerated, oprea2021h, alami2018unsupervised} is underperformance as the incomplete hand structure occluded by degradations. Besides that, the methods~\cite{mueller2018ganerated, oprea2021h} used to transfer synthetic hands to real hands are also unsatisfactory for our task.

\subsection{Ablation Study}
\label{sec44_ablation}
\begin{figure}[!t]
    \centering
    \includegraphics[width=\linewidth]{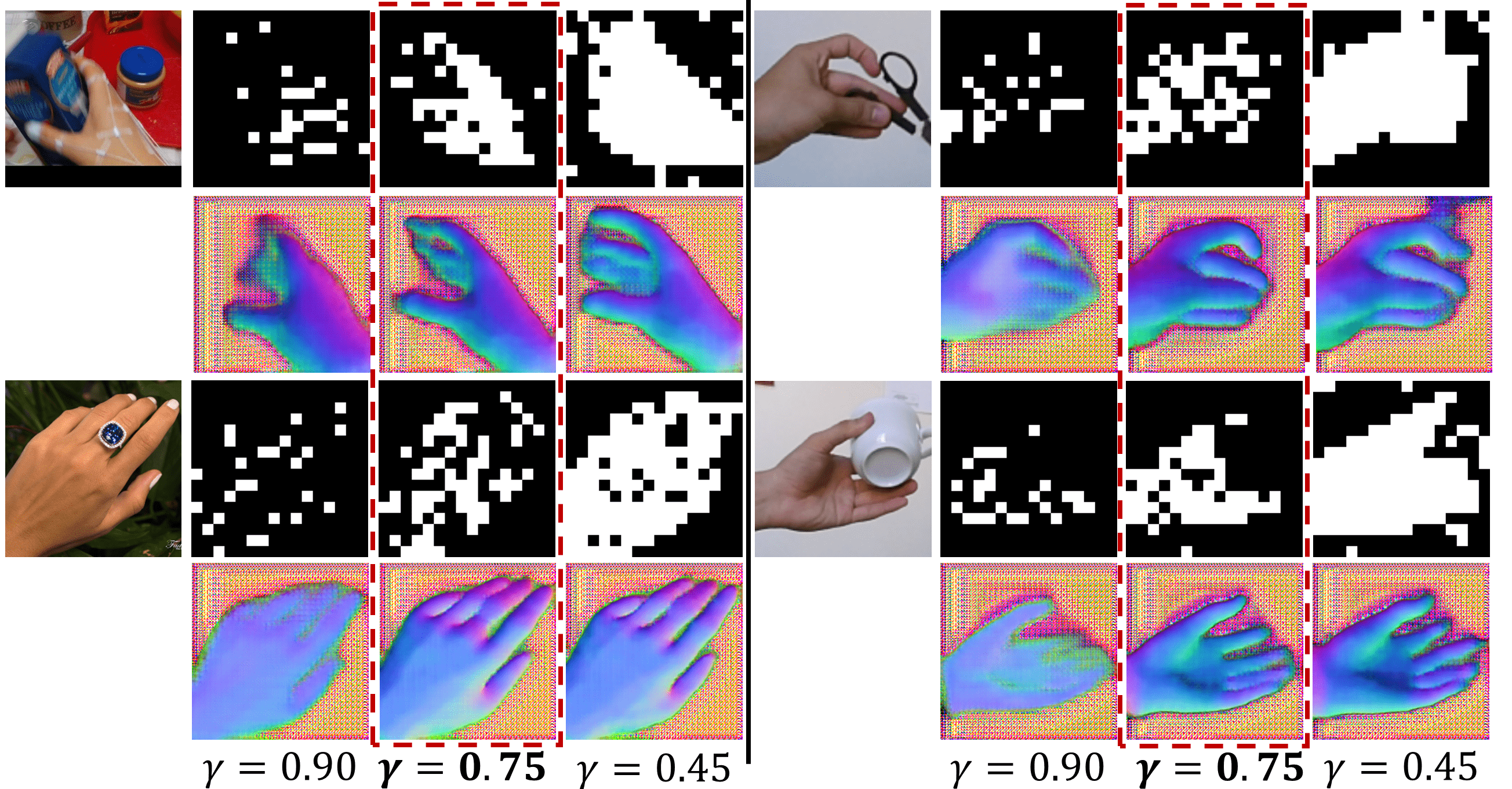}
    \vspace{-6mm}
    \caption{\textbf{Influence of $\gamma$ settings}. Testing results of the sketcher trained with different sampling masking-out ratios. }
    \vspace{-4mm}
    \label{fig10_vargamma}
\end{figure}

We first ablate the key components in the bare structure disentanglement: (i) \emph{Which module contributes more to the overall performance?} The quantitative of the two ablations are reported in Row-1 and Row-2 of \tablemk\ref{tab04_ablations}. ``w/ $S(X)$ w/o DAD'' is implemented by replacing DAD with unsupervised paradigms. The qualitative results in \figmk\ref{fig07_compareCyclegan} reveal the superiority of our DAD scheme. The qualitative results in \figmk\ref{fig04_ablationSX} show the guiding significance of $S(X)$ to other translation process. ``w/o $S(X)$ w/ DAD'' is implemented by feeding both inputs of our $\mathcal{G}$ with duplicate $X$. The above experiments indicate that the presence of $S(X)$ has a greater impact on our whole framework, and the design of DAD scheme make appearance wrapping more efficient. ~(ii) \emph{Is it necessary to construct a bare structure prior?} As shown in the last two rows of \figmk\ref{fig08_compareNormal}, a standardized domain without prior is powerless and susceptible to the background. In addition, as illustrated in \figmk\ref{fig08_compareNormal}, CNN specializes in local details (Row-3), while ViT excels in cross-region associations (Row-4). ~(iii) \emph{Why choose a normal map to represent the standardized domain?} We conduct variants with the other two easy-to-render candidates in \figmk\ref{fig02_structRep}. As shown in Row-3 and Row-4 of \tablemk\ref{tab04_ablations}, their testing performances are weaker than a normal one. The drawbacks of the IUV map include dependency on a fixed mesh
topology and dependency on a fixed mesh UV unwrapping. Consequently, compared to the normal map, the amount and diversity of available IUV data used to construct the bare structure prior become much smaller. (iii) \emph{How does the masking-out ratio affect the disentanglement performance?} As shown in \figmk\ref{fig10_vargamma}, a smaller one may introduce background distractions, while a larger one may inhibit the transmission of hand semantics. Coincidentally, $\gamma$\Equal$0.75$~\cite{he2022masked} fits our task well. (iv) The ablations on loss terms used in the sketcher's training are shown in Row-5 and Row-6 of \tablemk\ref{tab04_ablations}. It can be found that semi-supervised fine-tuning significantly improves the performance of our sketcher. Row-7 and Row-8 of \tablemk\ref{tab04_ablations} illustrate the effectiveness of our translator architecture and DAD scheme separately. 

\begin{table}[!t]
    \rowcolors{1}{}{lightgray}
    \begin{center}
        \resizebox{1\linewidth}{!}{
        \begin{tabular}{l|cc|cc}
        \noalign{\hrule height 1.5pt}
        Tasks 
        &\multicolumn{2}{c|}{$\mathbf{A}_{\text{1}} \rightarrow \mathbf{B}$}
        &\multicolumn{2}{c}{$\mathbf{A}_{\text{2}} \rightarrow \mathbf{B}$}\\
        \midrule
        Variants                    &$\text{FID}_{\text{v}}$ &$\text{KID}_{\text{v}}$ &$\text{FID}_{\text{v}}$ &$\text{KID}_{\text{v}}$ \\
        \midrule
        w/ $S(X)$ w/o DAD &$1213.29$      &$30.48\pm1.201$     &$935.78$          &$13.77\pm0.435$        \\
        w/o $S(X)$ w/ DAD &$1259.33$      &$31.05\pm1.569$     &$962.51$          &$14.13\pm0.647$        \\
        \midrule
        w/ $S_{uv} (c=2)$    &$1285.37$      &$31.49\pm2.172$        &$702.90$          &$16.68\pm2.297$        \\
        w/ $S_{d} ~(c=1)$    &$1306.29$      &$33.82\pm2.192$        &$725.95$          &$17.01\pm2.031$        \\
        \midrule
        w/o $L_{\mathrm{ad}}$(\equationmk\ref{eqn_tokenloss})             &$1172.12$      &$32.07\pm0.112$        &$712.15$          &$16.73\pm1.032$        \\
        w/o $L_{\mathcal{E}_s, \mathcal{F}_s}$(\equationmk\ref{eqn_ef2}) &$1161.14$      &$30.61\pm0.178$        &$694.73$          &$16.32\pm0.893$        \\
        \midrule
        w/ U-net $\mathcal{G}$               &$1174.90$      &$29.87\pm0.201$        &$685.50$          &$16.07\pm1.034$        \\
        \midrule
        w/ Syn-Pix2pix                      &$1080.33$      &$29.09\pm0.195$        &$680.32$          &$15.99\pm1.132$        \\
        \midrule
        Full                                &$\bm{994.67}$ &$\bm{28.67\pm0.916}$   &$\bm{673.43}$     &$\bm{15.72\pm1.209}$    \\
        \noalign{\hrule height 1.5pt}
        \end{tabular}
    }
    \end{center}
    \vspace{-4mm}
    \caption{\textbf{Ablation studies on $\mathbf{A}_{\text{1}} \rightarrow \mathbf{B}$ and $\mathbf{A}_{\text{2}} \rightarrow \mathbf{B}$.} The first column describes the ablation item, and the other columns show the impact of the ablation on the final translation result. }
    \label{tab04_ablations}
    \vspace{-4mm}
\end{table}
\section{Conclusion} 
This work pioneers a semi-supervised image-to-image translation to recover the hand appearance that was originally degraded during the marker-based \mocap process. Since this task also implies the degradation in hand structure, a prior-based sketcher is first proposed to disentangle the bare hand structure map from images. Later, an efficient adversarial scheme is devised to guide the translator to selectively wrap the appearance from the original image to the above structure map. This framework enables data from marker-based \mocap to regain complete and photo-realistic hand appearance. It also provides a novel avenue to the dilemma in the simultaneous acquisition of hand appearance and motion data. 

\noindent\textbf{Limitations and future work}. Our framework may become unstable when the input is severely degraded. Although only the single-hand case is verified, this prior-based method could also be adapted to multi-hand or body applications. In addition, improving it to tackle sequential problems could bring more benefits to the community.
\clearpage
\begingroup

\twocolumn[
\begin{center}
    {\Large \bf \Large{Stability-driven Contact Reconstruction From Monocular Color Images} \\ -- Supplementary Material -- \par}
  \vspace*{30pt}
\end{center}
]
\appendix

\setcounter{table}{0}
\setcounter{figure}{0}
\setcounter{equation}{0}
\renewcommand{\thetable}{\thesection.\arabic{table}}
\renewcommand{\thefigure}{\thesection.\arabic{figure}}
\renewcommand{\theequation}{\thesection.\arabic{equation}}


\section{Overview}
In this supplementary document, we first introduce the hand saliency network to assist our ViT sketcher's training in \secmk\ref{01_saliency}. Then, we explain the way to construct the partner domain to assist our dual adversarial discrimination (DAD) scheme in \secmk\ref{02_partner}. After that, we describe the data augmentation strategies used in each piece of training in \secmk\ref{03_augment}. We further add more experimental results of our methods (\secmk\ref{04_exp}), as well as the discussions about its failure cases (\secmk\ref{05_failure}). They were not included in the main paper due to the page limit.


\section{Hand Saliency Estimation}
\label{01_saliency}
\noindent\textbf{Utilization}. Our estimator regresses the visible hand saliency $M(X) \in [0,1]^{(h,w)}$ from a hand-centered image $X \in \mathbb{R}^{(3, h,w)}$. Because $S$ is also an image domain containing bare hand structure, the estimator is also compatible with regressing $M(S)$ from $S$. Compared with those generic instance segmentation approaches~\cite{bolya2019yolact, peng2020deep}, $M(X)$ may be imperfect, but it is sufficient to reduce the effect of the background. Furthermore, as shown in \figmk\ref{fig04_totalmap}, $M[S(X)]$ is more valuable for our task, which retains the complete bare hand structure defined in $S$. ~~In our framework, $M(X)$ plays the following roles: (i)  $\bm{m}(X)^\star = \mathrm{MaxPool}(M(X), p)$ is used as a teacher to provide patch-wise saliency supervision during training of the sketcher's MLP. (ii) $M[S(X)]$ is used as a mask to compute MSE only for the background part during the training of the translator. (iii) $M[S(X)]$ is utilized as a mask to extract domain features only for the non-background part within DNN perceptual metrics.

\noindent\textbf{Architecture}. As illustrated in \figmk\ref{fig01_handsaliency}, a header composed of two parallel convolution layers is first used to extract features from input images. Subsequently, those two branches are concatenated and fed into an encoder. The encoder is composed of 5 residual blocks, each of them combined with 2D convolution layers and rectified linear unit functions. As a result, receptive fields with gradual enlargement are obtained. The decoder adopts a symmetric structure similar to the encoder that also consists of 5 stacked residual blocks but with up-sampling behind each block. For the first decoder layer, after scaling up by up-sampling, the feature map produced by the last encoder layer was concatenated with the later encoder layer. Similarly, other decoder layers do the same up-sampling and concatenate with the encoder output with the same resolution. Except for the last layer, leaky-ReLU is adopted for activation. Finally, a hand saliency with 256 $\times$ 256 is estimated.

\begin{figure}[h]
    \centering
    \includegraphics[width=\linewidth]{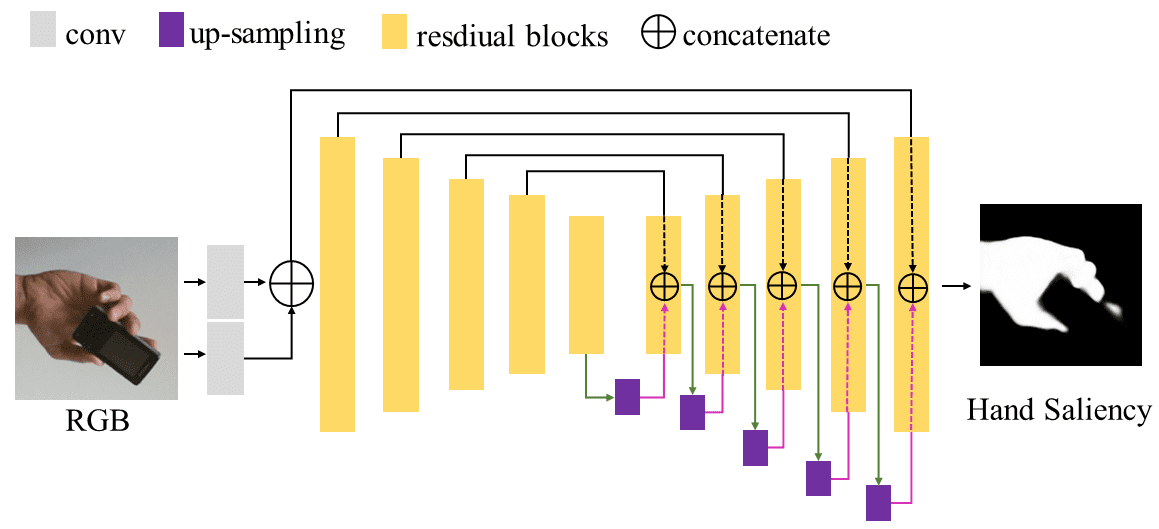}
    \caption{\textbf{Hand saliency estimation network}. We use an encoder-decoder architecture network to estimate hand saliency. We do the skip connection between the encoder and decoder with the same resolution.}
    \vspace{-4mm}
    \label{fig01_handsaliency}
\end{figure}
\section{Partner Domain Construction}
\label{02_partner}
The partner domain $\tilde{\mathbf{B}}$ provides a bridge for unpaired translation from source domain $\mathbf{A}$ to target domain $\mathbf{B}$ according to our DAD scheme. During the training of our translator, $X_{\tilde{\mathbf{B}}} \in \tilde{\mathbf{B}}$ is augmented by $X_\mathbf{B} \in \mathbf{B}$ as follows: 
\begin{equation}
    \begin{aligned}
        X_{\tilde{\mathbf{B}}} &= \tilde{N}(X_\mathbf{B}) \\
        &= (1- M[S(X)]) \odot X_\mathbf{B} + M[S(X)] \odot D(X_\mathbf{B})
    \end{aligned}
    \label{eqn_semiS}
\end{equation}
where $M[S(X)]$ ensures that the degradations only occur in the hand region. $D(X_\mathbf{B})$ is used to simulate various markers, gloves, and objects. In practice, we create a variety of hand-specific degradations: (i) Spot degradations centered at 
visible hand joints. The locations are estimated by the off-the-shelf 2D key-point estimator~\cite{wang2019srhandnet}. (ii) Line degradations distributed along visible hand affinities~\cite{cao2017realtime}, \thatis between adjacent visible joints. (iii) Region degradations approximated by randomly enclosing polygons in $M[S(X)]$; (iv) Whole degradation in $M[S(X)]$. The number and color of each degradation type are random. They are interpolated randomly with the original pixel value. Some examples are shown in \figmk\ref{fig03_dataaugment}. 
\section{Training Data Augmentation}
\label{03_augment}

\noindent\textbf{Image domain}. During the training of the MLP, discrete VAE$\{\mathcal{T}_{s}, \mathcal{F}_{s}\}$, attention decoder $\mathcal{E}_{s}$, and translator $\mathcal{G}$, the following data enhancement methods are adopted to augment the input RGB: (i) flip randomly up-down and left-right; (ii) rotate $\theta \in U(-\pi, \pi)$ randomly centered on the hand area; (iii) scale $s \in U(0.8, 1.2)$ randomly centered on the hand area; (iv) blur randomly with a kernel size $k \in U(3, 9)$. 

\noindent\textbf{Structure domain}. The structure map represented as depth $S_d$ or IUV $S_{uv}$ used in ablations is augmented in the same way as the RGB image. For normal map $S_n$, because the pixel with the 2D coordinate $(u,v)$ records the normal direction of the hand surface point in the camera coordinate system, it also changes after a flip or rotation augmentation: 
\begin{equation}
    \begin{aligned}
        S_n (u,v)' = R(\theta) \cdot S_n (u,v)
    \end{aligned}
    \label{eqn_augN}
\end{equation}
where $\theta$ is the accumulated angle from flip and rotation. Other forms of augmentation do not cause changes in the normal vector. 
\section{More Experiments}
\label{04_exp}

\noindent\textbf{Human perceptual metrics.} The human perceptual survey about the translation authenticity of the results from different frameworks is completed in Amazon Mechanical Turk (AMT). At the beginning of the questionnaire, participants were instructed to select one ``real'' candidate in each question that best matches the appearance of the bare hand and is most consistent with the semantics in the source image. We collect 2K questionnaires and use the percentage of each method's score divided by the total number of people (2K) as the evaluation of the translation quality of each method, with higher scores representing better results. As shown in \tablemk\ref{tab02_compareAMT}, our framework obtains the majority of votes for best translating from both $\mathbf{A}_{\text{1}} \rightarrow \mathbf{B}$ and $\mathbf{A}_{\text{2}} \rightarrow \mathbf{B}$. It is worth noting that not all participants evaluated all six methods due to the random assignment process. However, the number of participants in each method evaluated ranged from 55\% to 60\% of the total number of participants, for this reason, our numbers may be different from the original baselines. The translation quality of CUT~\cite{park2020contrastive} is second, and several other methods~\cite{zhu2017unpaired, mueller2018ganerated, oprea2021h, alami2018unsupervised} based on cycle consistency have poorer quality in our task. 

\begin{table}[!t]
    \rowcolors{1}{}{lightgray}
    \begin{center}
        \resizebox{1\linewidth}{!}{
    \begin{tabular}{c|c|c}
    \noalign{\hrule height 1.5pt}
    Tasks 
    &\multicolumn{1}{c|}{$\mathbf{A}_{\text{1}} \rightarrow \mathbf{B} \uparrow$} 
    &\multicolumn{1}{c}{$\mathbf{A}_{\text{2}} \rightarrow \mathbf{B} \uparrow$}\\
    \midrule
    CycleGAN~\cite{zhu2017unpaired}   
    &$18.46\%\pm0.9\%$                              &$22.59\%\pm1.7\%$              \\
    GANerated~\cite{mueller2018ganerated}
    &$14.51\%\pm2.3\%$                              &$17.47\%\pm2.0\%$              \\
    H-GAN~\cite{oprea2021h}
    &$11.12\%\pm1.7\%$                              &$9.91\%\pm2.8\%$              \\
    UAG~\cite{alami2018unsupervised}
    &$13.93\%\pm0.4\%$                              &$12.59\%\pm1.2\%$              \\
    CUT~\cite{park2020contrastive}
    &$21.26\%\pm1.6\%$                              &$23.19\%\pm3.2\%$                \\
    Ours
    &$\bm{28.16\%\pm1.1\%}$                         &$\bm{32.37\%\pm2.3\%}$                  \\
    \noalign{\hrule height 1.5pt}
    \end{tabular}
    }
    \end{center}
    \vspace{-4mm}
    \caption{\textbf{AMT perceptual evaluation.} AMT \emph{real vs fake} test on $\mathbf{A}_{\text{1}} \rightarrow \mathbf{B}$ and $\mathbf{A}_{\text{2}} \rightarrow \mathbf{B}$.}
    \label{tab02_compareAMT}
\end{table}

\noindent\textbf{Improvement to pose estimation.} We quantitatively evaluated the effect of recovering the bare appearance on the accuracy of hand pose estimation. Two different 2D pose estimators~\cite{cao2017realtime, wang2019srhandnet} are adopted to estimate the accuracy of key-points localization before and after restoring the bare appearance for data from two datasets~\cite{garcia2018first, zimmermann2019freihand}. As shown in \tablemk\ref{tab03_compareHPE}, the accuracy of the pose estimators is generally improved on the datasets with recovered appearance. This is one of the most direct ways our framework can contribute to downstream tasks. 

\begin{table}[!t]
    \rowcolors{1}{}{lightgray}
    \begin{center}
        \resizebox{1\linewidth}{!}{
        \begin{tabular}{l|cc|cc}
        \noalign{\hrule height 1.5pt}
        Estimator 
        &\multicolumn{2}{c|}{Openpose~\cite{cao2017realtime}}
        & \multicolumn{2}{c}{SRNet~\cite{wang2019srhandnet}} \\
        \midrule
        Dataset Version                 &Original &Translated &Original &Translated \\
        \midrule
        FPHAB~\cite{garcia2018first}      &$0.81$ &$0.85$        &$0.86$   &$0.91$  \\
        FreiHand~\cite{zimmermann2019freihand}   &$0.87$ &$0.92$      &$0.89$  &$0.93$  \\
        \noalign{\hrule height 1.5pt}
        \end{tabular}
    }
    \end{center}
    \vspace{-4mm}
    \caption{\textbf{Hand pose estimation performance} on the original datasets and their appearance recovery version translated by our framework. PCK0.2 score is adopted as the accuracy criterion.}
    \label{tab03_compareHPE}
\end{table}

\noindent\textbf{Universality of structure domain.} In our framework, a bare structure prior defined on a standardized domain is built explicitly. Interestingly, in \splicevit\cite{tumanyan2022splicing}, a similar domain is constructed in its implicit appearance wrapping process. As shown in \figmk\ref{fig02_compare_vit}, in the first few iterations, the translator in \splicevit tends to translate the structural reference to a uniform domain that contains only visible structure information. This finding indirectly confirms that the design of our framework to disentangle the bare structure is reasonable and efficient. 

\begin{figure}[!t]
    \centering
    \includegraphics[width=\linewidth]{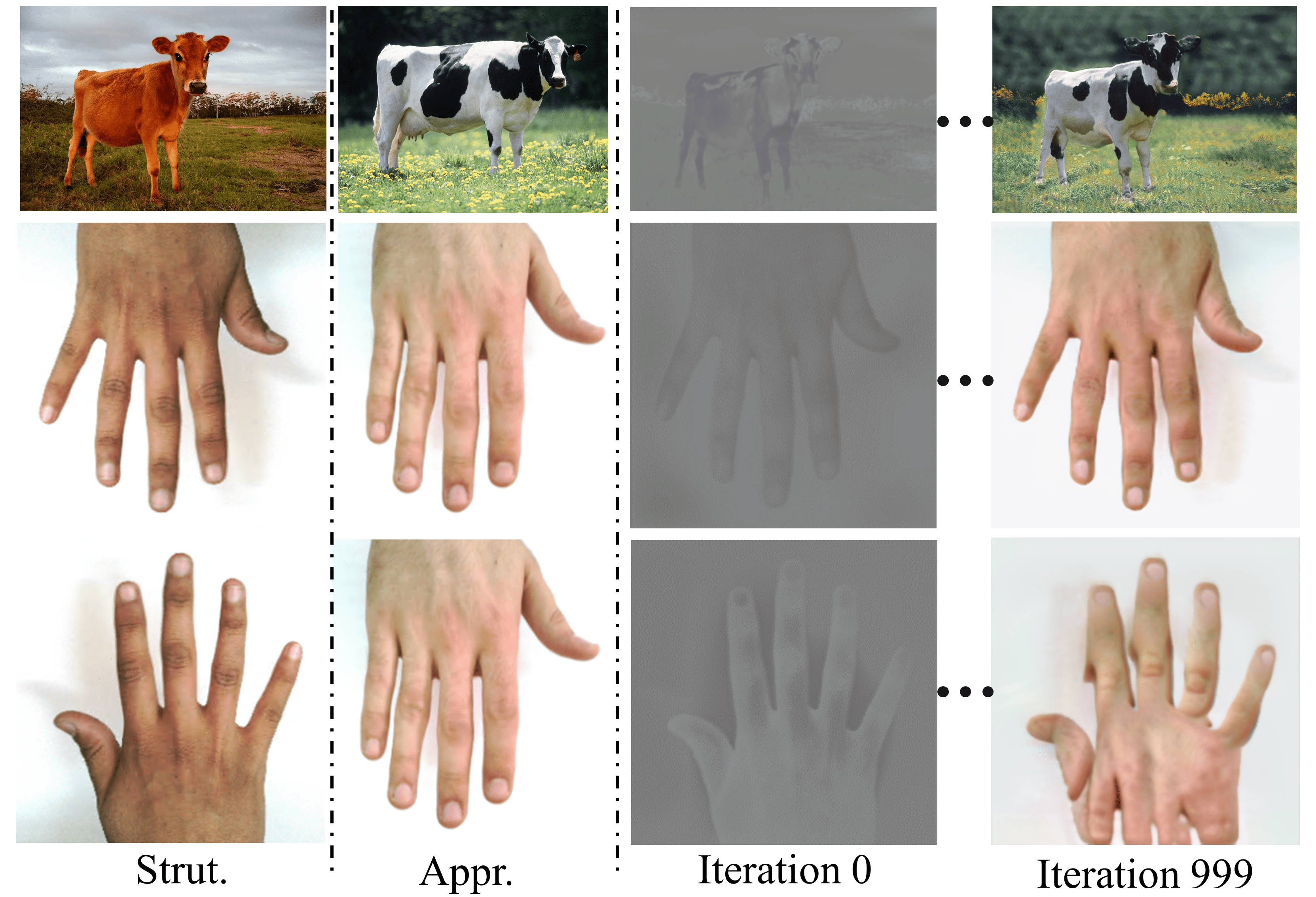}
    \caption{\textbf{\splicevit wrapping process}. From left to right: the structure reference, the appearance reference, the wrapping results in the first few iterations, and the final wrapping results. }
    \vspace{-4mm}
    \label{fig02_compare_vit}
\end{figure}

\noindent\textbf{More recovery results.} We show more qualitative results for our framework in hand appearance recovery from $\mathbf{A}_{\text{1}} \rightarrow \mathbf{B}$ in \figmk\ref{fig03_fullpipeline_res0} and $\mathbf{A}_{\text{2}} \rightarrow \mathbf{B}$ in \figmk\ref{fig03_fullpipeline_res1}. The sampled hand region $[M](X)$ and the disentangled structure map $S(X)$ are also comprehensively illustrated for each example. 

An interesting phenomenon is that during the translation $\mathbf{A}_{\text{2}} \rightarrow \mathbf{B}$, the object may be partially removed (the part that occludes the hand) or completely removed. This may be because we do not assign as much weight to the constrained background-consistent MSE loss in training as pix2pix~\cite{wang2018high}. In this way, the object part outside the bare hand region may also be penalized by the result discriminator $\mathcal{D}_\mathbf{B}^{(r)}$ in our DAD scheme, which makes the translator tend to erase them. 
\section{Failure Cases}
\label{05_failure}
As shown in \figmk\ref{fig12_limit}, our framework is unstable when the input is severely degraded. In this case, hand features come entirely from the illusions of our CNN translator, which has anisotropic convolution kernels.

\begin{figure}[!t]
    \centering
    \includegraphics[width=\linewidth]{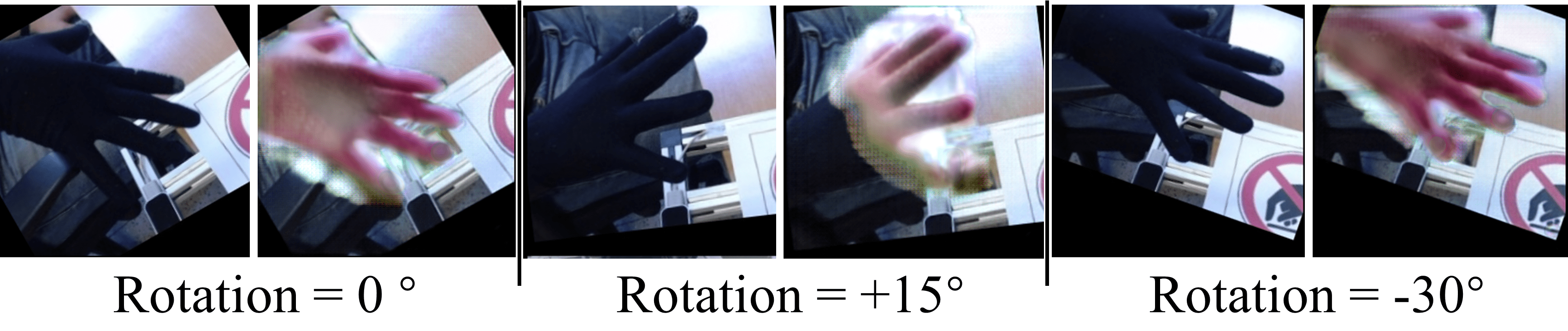}
    \caption{\textbf{Severe degradation cases}. When the input appearance is severely degraded, our model is not rotation-invariant. }
    \vspace{-4mm}
    \label{fig12_limit}
\end{figure}


\begin{figure}[!t]
    \centering
    \includegraphics[width=\linewidth]{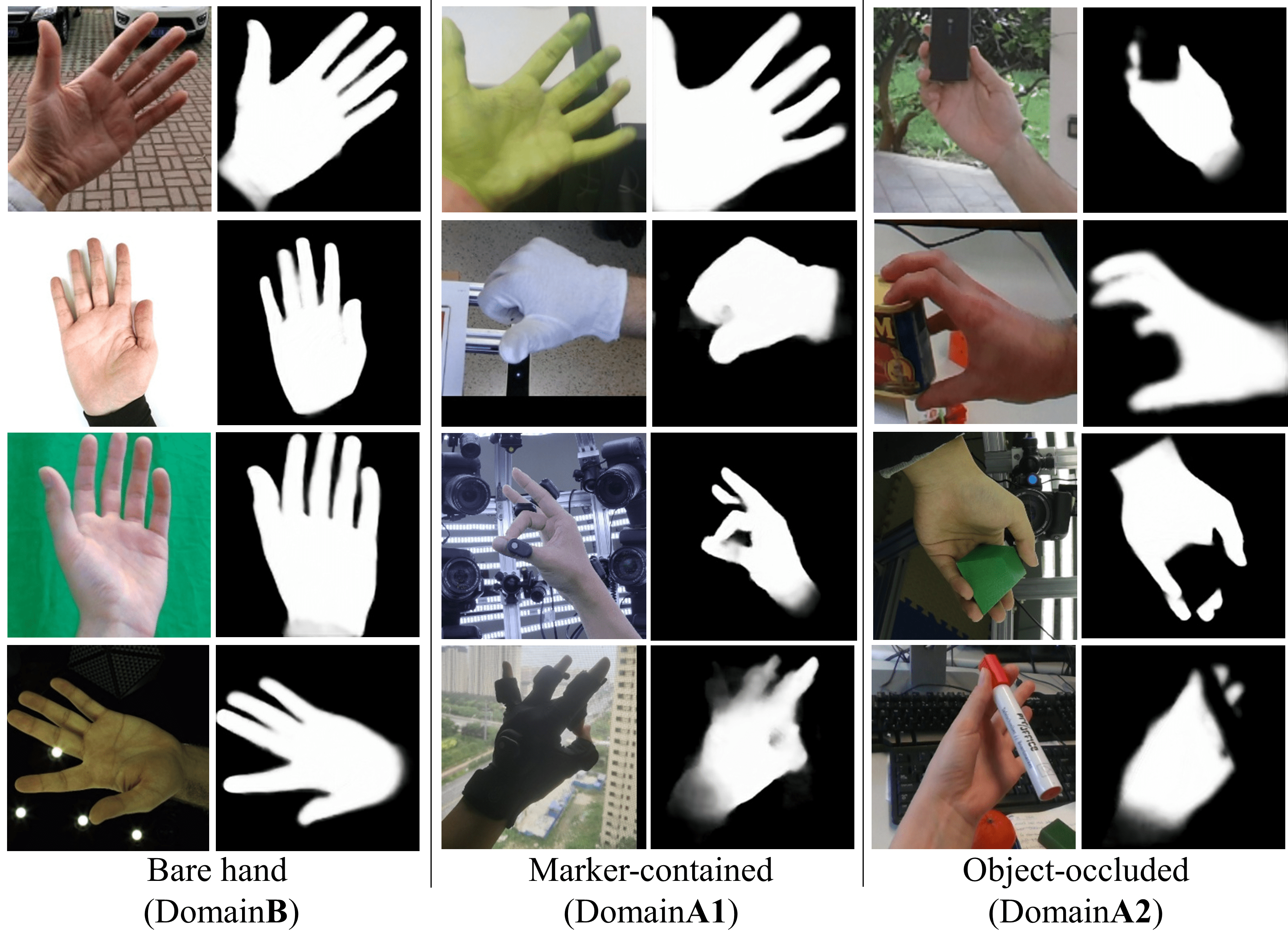}
    \caption{\textbf{Additional hand saliency results on different domains}. From left to right: hand saliency estimations on domain $\mathbf{B}$, hand saliency estimations on domain $\mathbf{A}_{\text{1}}$ and hand saliency estimations on domain $\mathbf{A}_{\text{2}}$.}
    \vspace{-4mm}
    \label{fig02_handsaliency_result}
\end{figure}

\begin{figure}[!t]
    \centering
    \includegraphics[width=\linewidth]{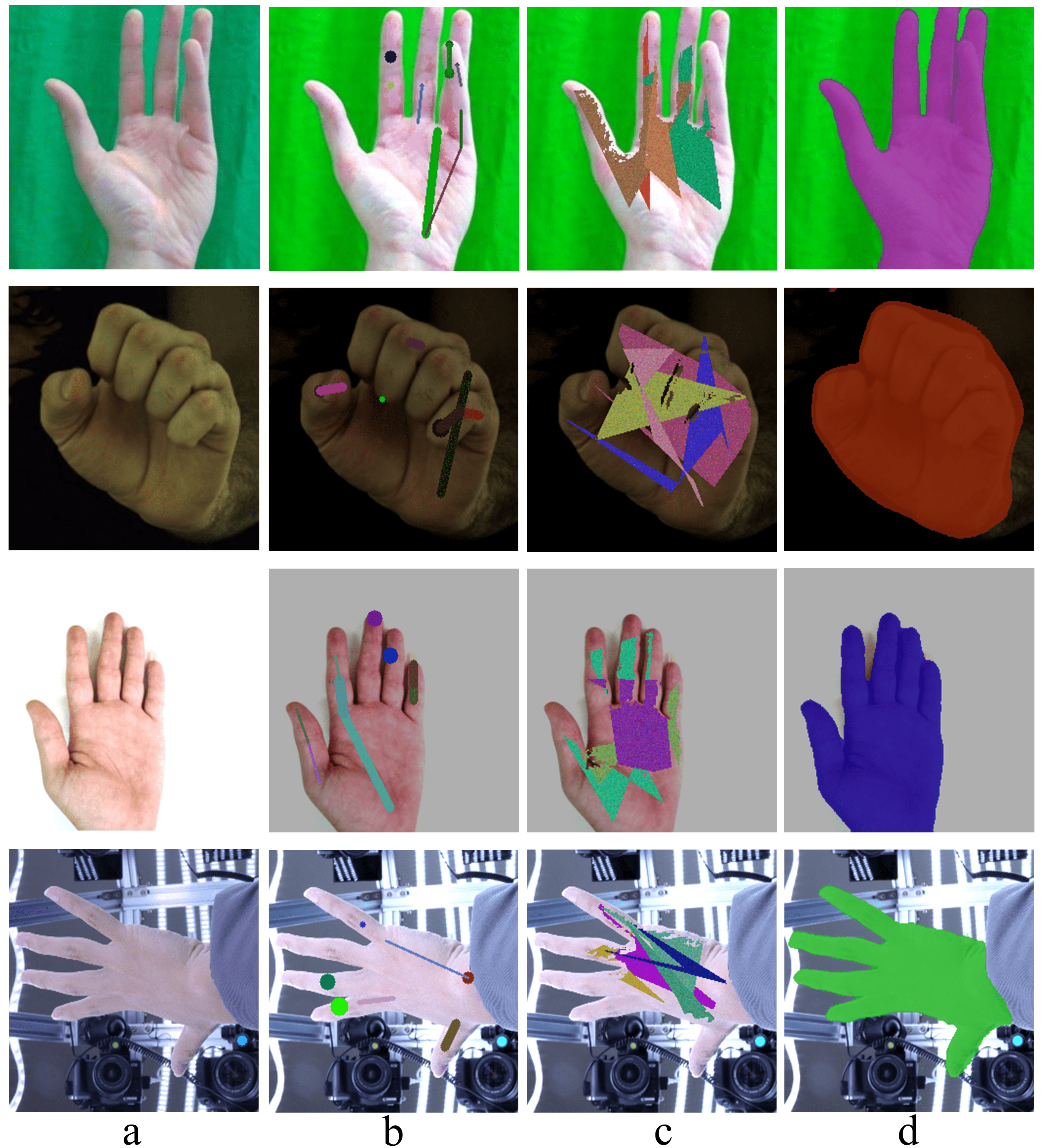}
    \caption{\textbf{Degradation process to obtain partner domain}. From left to right: $\mathbf{a.}$Input images; $\mathbf{b.}$Combined with spot and line degradations based on visible hand joints and affinities. $\mathbf{c.}$Polygon-based region degradations. $\mathbf{d.}$Mask-based whole degradations.}
    \vspace{-4mm}
    \label{fig03_dataaugment}
\end{figure}

\begin{figure*}[!t]
    \centering
    \includegraphics[width=\linewidth]{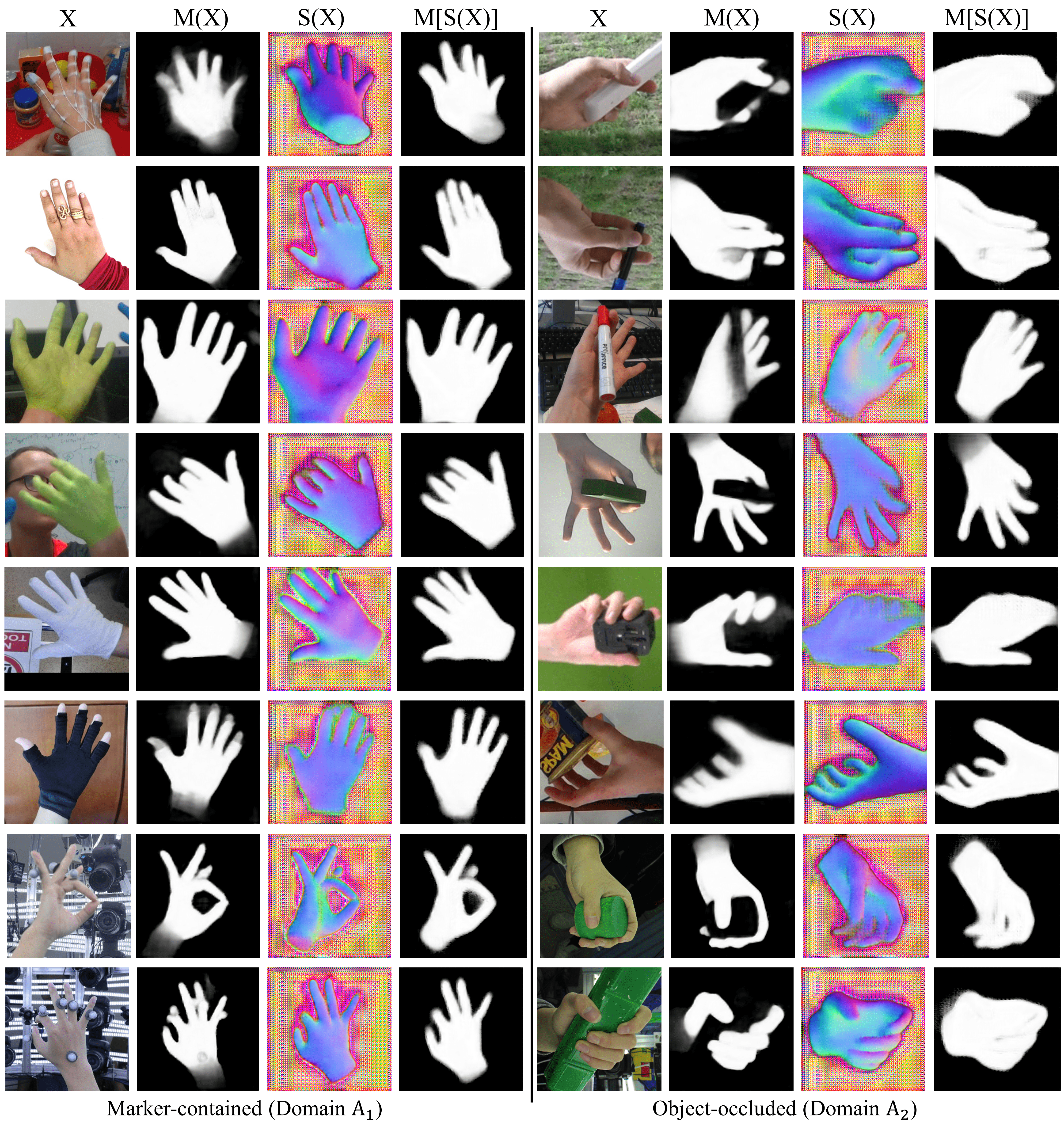}
    \caption{\textbf{Saliency Estimations from images and structure maps}. Our saliency estimator can both estimate visible hand saliency $M(X)$ from an RGB image $X$, or hand structure saliency $M[S(X)]$ from a structure map $S(X)$. }
    \vspace{-4mm}
    \label{fig04_totalmap}
\end{figure*}

\begin{figure*}[!t]
    \centering
    \includegraphics[width=\linewidth]{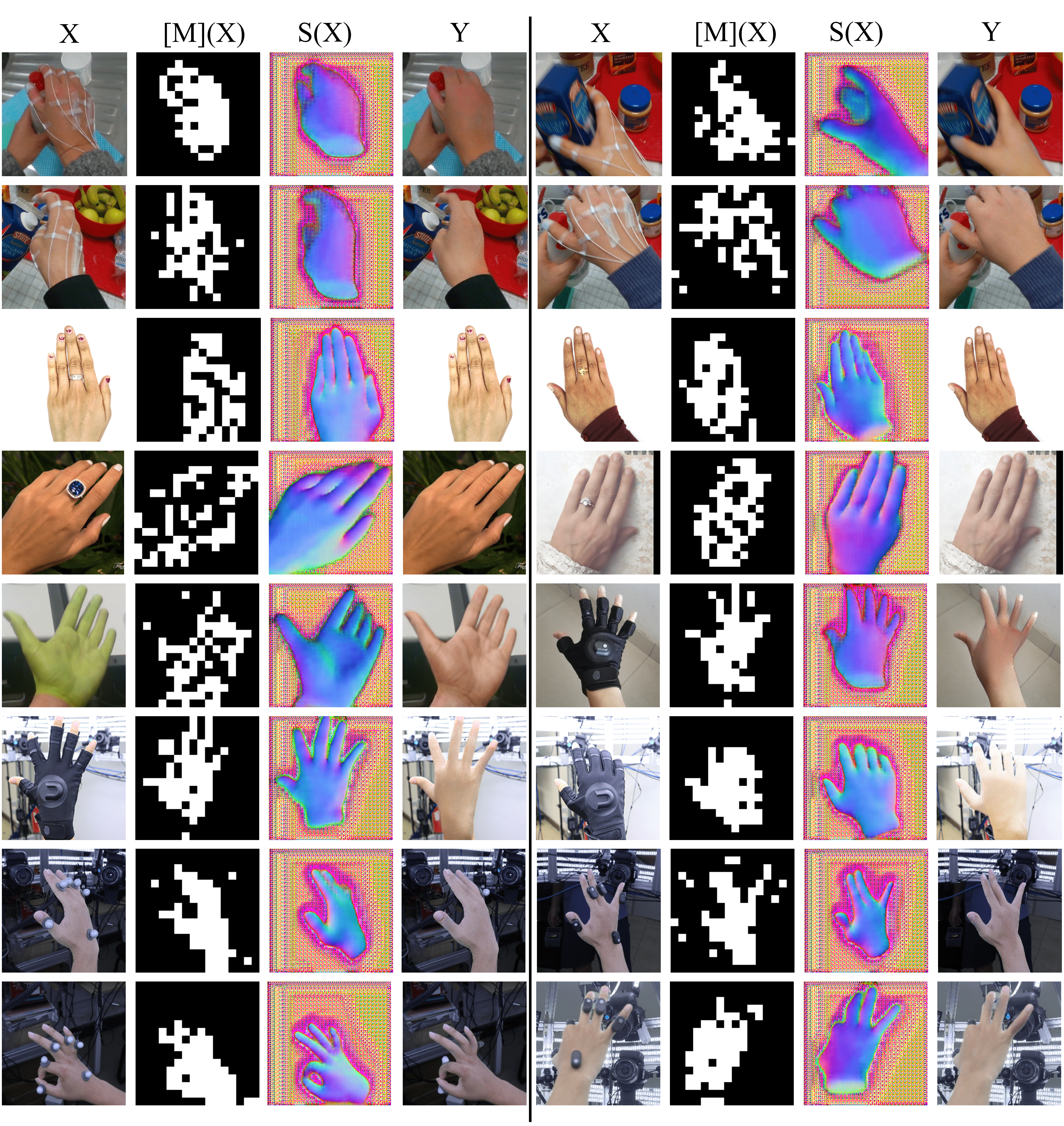}
    \caption{\textbf{Translation results of our framework on $\mathbf{A}_{\text{1}} \rightarrow \mathbf{B}$.} Two groups of results are presented in each row. Each group of results includes the input image $X$, the sampled mask $[M](X)$, the disentangled structure map $S(X)$ and the appearance wrapping result $Y$.}
    \vspace{-5mm}
    \label{fig03_fullpipeline_res0}
\end{figure*}

\begin{figure*}[!t]
    \centering
    \includegraphics[width=\linewidth]{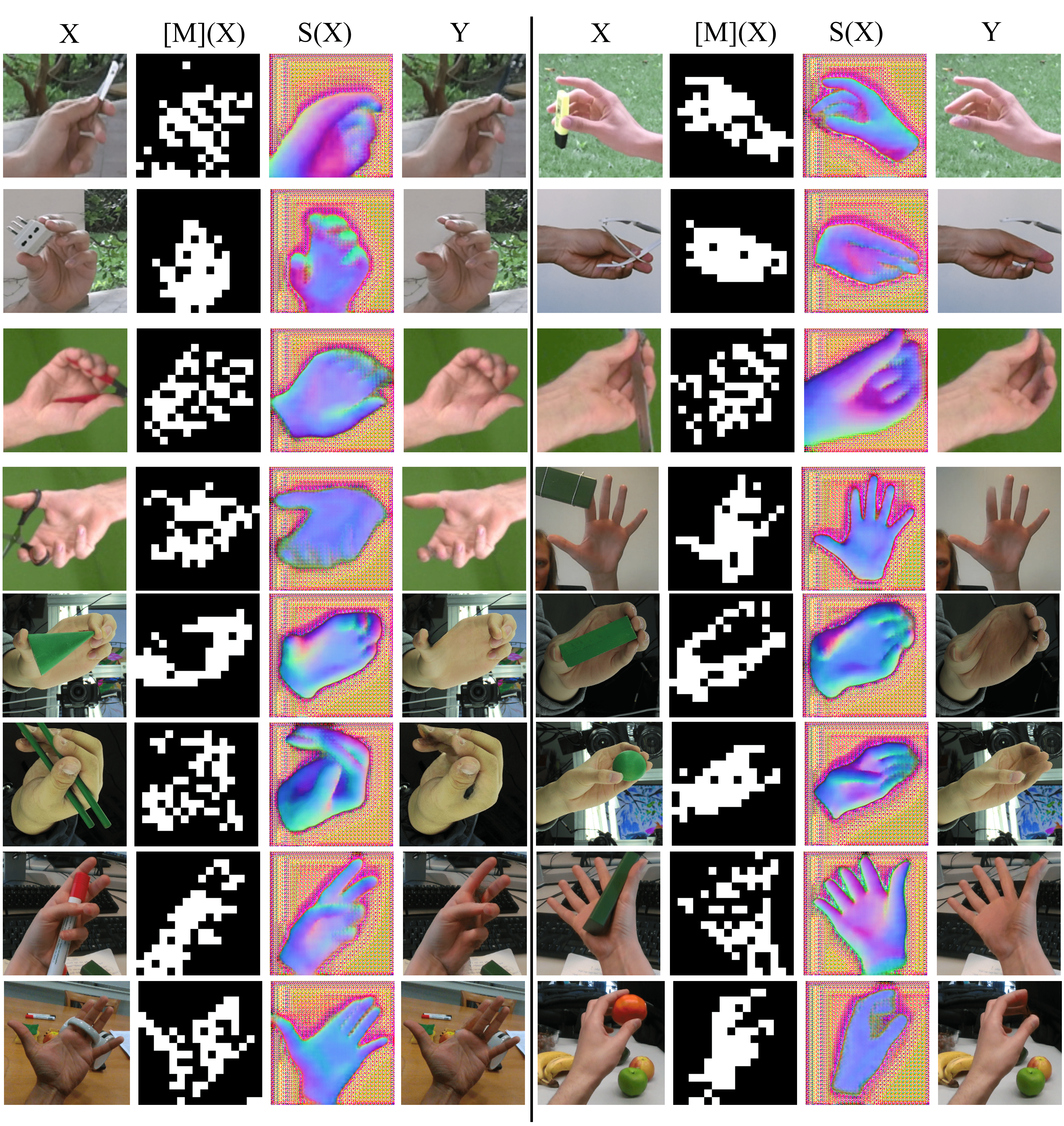}
    \caption{\textbf{Translation results of our framework on $\mathbf{A}_{\text{2}} \rightarrow \mathbf{B}$.} Two groups of results are presented in each row. Each group of results includes the input image $X$, the sampled mask $[M](X)$, the disentangled structure map $S(X)$ and the appearance wrapping result $Y$.}
    \vspace{-5mm}
    \label{fig03_fullpipeline_res1}
\end{figure*}


{\small
\bibliographystyle{ieee_fullname}
\bibliography{egbib}
}

\end{document}